\documentclass[10pt,twocolumn,letterpaper]{article}

\usepackage{cvpr}
\usepackage{times}
\usepackage{epsfig}
\usepackage{graphicx}
\usepackage{amsmath}
\usepackage{amssymb}
\usepackage{algorithm}
\usepackage[noend]{algorithmic}
\usepackage{microtype}
\usepackage{comment}
\usepackage{tabularx}
\usepackage{multirow}
\usepackage{rotating}
\newcolumntype{Y}{>{\centering\arraybackslash}X}
\usepackage[export]{adjustbox}

\usepackage{enumitem}
\setitemize[0]{leftmargin=15pt}

\newenvironment{tight_itemize}{
\begin{itemize}[leftmargin=15pt]
  \setlength{\topsep}{0pt}
  \setlength{\itemsep}{0pt}
  \setlength{\parskip}{0pt}
  \setlength{\parsep}{0pt}
}{\end{itemize}}

\newcommand{\ZL}[1]{\textcolor{black}{{#1}}}

\usepackage[pagebackref=true,breaklinks=true,letterpaper=true,colorlinks,bookmarks=false]{hyperref}

\cvprfinalcopy % *** Uncomment this line for the final submission

\def\cvprPaperID{4402} % *** Enter the CVPR Paper ID here

\ifcvprfinal\fi
\begin{document}

%%%%%%%%% TITLE
\title{Through the Looking Glass: Neural 3D Reconstruction of Transparent Shapes}

\author{{Zhengqin Li}\thanks{These two authors contributed equally} \quad {Yu-Ying Yeh}\textsuperscript{*} \quad {Manmohan Chandraker}\\
{University of California, San Diego}\\
{\tt\small { {\{zhl378, yuyeh, mkchandraker\}}@eng.ucsd.edu}}
\vspace{-2mm}
}

\maketitle

\begin{abstract}
Recovering the 3D shape of transparent objects using a small number of unconstrained natural images is an ill-posed problem. Complex light paths induced by refraction and reflection have prevented both traditional and deep multiview stereo from solving this challenge. We propose a physically-based network to recover 3D shape of transparent objects using a few images acquired with a mobile phone camera, under a known but arbitrary environment map. Our novel contributions include a normal representation that enables the network to model complex light transport through local computation, a rendering layer that models refractions and reflections, a cost volume specifically designed for normal refinement of transparent shapes and a feature mapping based on predicted normals for 3D point cloud reconstruction. We render a synthetic dataset to encourage the model to learn refractive light transport across different views. Our experiments show successful recovery of high-quality 3D geometry for complex transparent shapes using as few as 5-12 natural images. \href{https://github.com/lzqsd/TransparentShapeReconstruction.git}{Code and data} are publicly released.

\end{abstract}

\vspace{-0.2cm}
\section{Introduction}
\label{sec:intro}
\vspace{-0.2cm}

Transparent objects abound in real-world environments, thus, their reconstruction from images has several applications such as 3D modeling and augmented reality. However, their visual appearance is far more complex than that of opaque objects, due to complex light paths with both refractions and reflections. This makes image-based reconstruction of transparent objects extremely ill-posed, since only highly convoluted intensities of an environment map are observed. In this paper, we propose that data-driven priors learned by a deep network that models the physical basis of image formation can solve the problem of transparent shape reconstruction using a few natural images acquired with a commodity mobile phone camera.

\begin{figure}[t]
\centering
\includegraphics[width=3.3in]{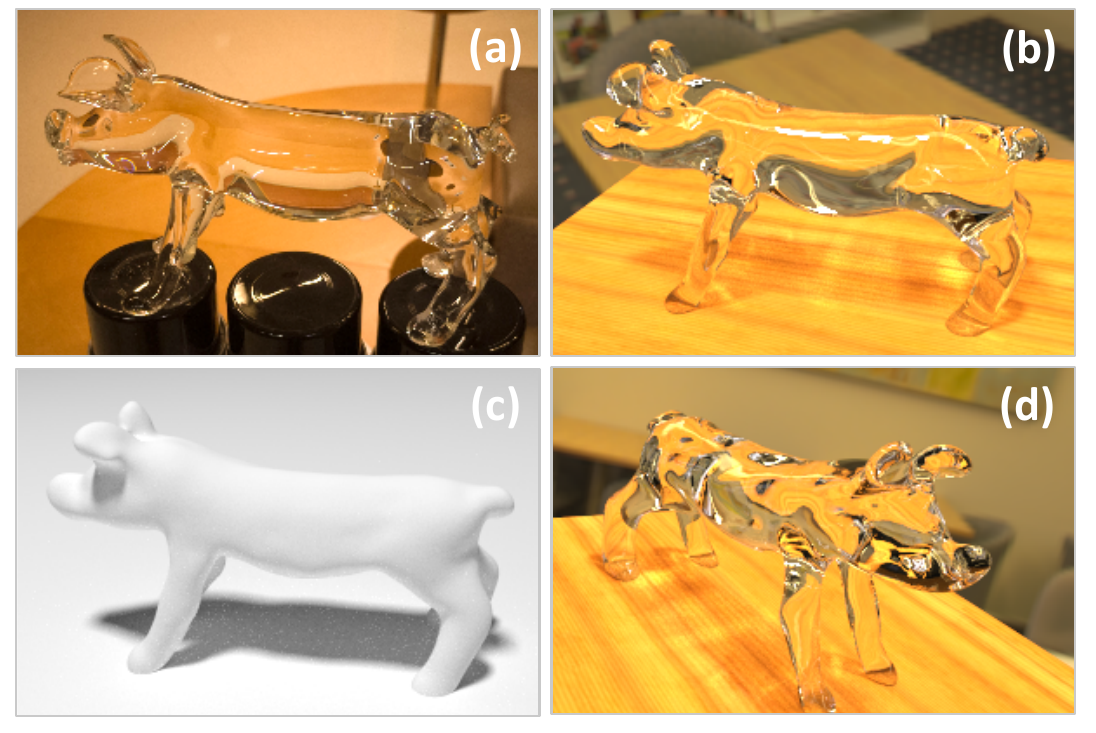}
\caption{We present a novel physically-based deep network for image-based reconstruction of transparent objects with a small number of views. (a) An input photograph of a real transparent object captured under unconstrained conditions (1 of 10 images). (b) and (c): The reconstructed shape rendered under the same view with transparent and white diffuse material. (d) The reconstructed shape rendered under a novel view and environment map. }
\label{fig:teaser}
\vspace{-0.3cm}
\end{figure}

\begin{figure*}[t]
\centering
\includegraphics[width=6.6in]{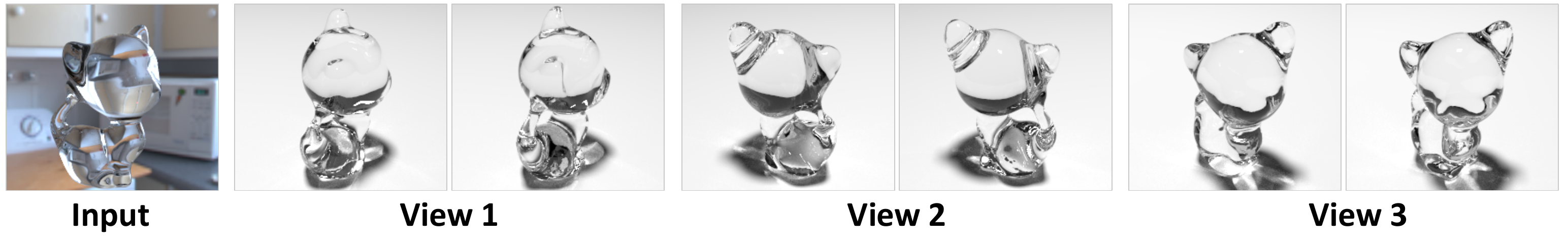}
\vspace{-0.2cm}
\caption{Reconstruction using 10 images of synthetic \emph{kitten} model. The left image is rendered with the reconstructed shape while the right image is rendered with the ground-truth shape.}
\label{fig:renderedKitten}
\vspace{-0.3cm}
\end{figure*}

While physically-based networks have been proposed to solve inverse problems for opaque objects \cite{li2018svbrdf},
%where image formation is governed by a bidirectional reflectance distribution function \cite{li2018svbrdf}. 
the complexity of light paths is higher for transparent shapes and small changes in shape can manifest as severely non-local changes in appearance. However, the physical basis of image formation for transparent objects is well-known -- refraction at the interface is governed by Snell's law, the relative fraction of reflection is determined by Fresnel's equations and total internal reflection occurs when the angle of incidence at the interface to a medium with lower refractive index is below critical angle. These properties have been used to delineate theoretical conditions on reconstruction of transparent shapes \cite{lightpath}, as well as acquire high-quality shapes under controlled settings \cite{wu2018full,yeungetal}. In contrast, we propose to leverage this knowledge of image formation within a deep network to reconstruct transparent shapes using relatively unconstrained images under arbitrary environment maps.

Specifically, we use a small number of views of a glass object with known refractive index, observed under a known but arbitrary environment map, using a mobile phone camera. Note that this is a significantly less restricted setting compared to most prior works that require dark room environments, projector-camera setups or controlled acquisition of a large number of images. Starting with a visual hull construction, we propose a novel in-network differentiable rendering layer that models refractive light paths up to two bounces to refine surface normals corresponding to a backprojected ray at both the front and back of the object, along with a mask to identify regions where total internal reflection occurs. Next, we propose a novel cost volume to further leverage correspondence between the input image and environment map, but with special considerations since the two sets of normal maps span a four-dimensional space, which makes conventional cost volumes from multiview stereo intractable. Using our \ZL{differentiable rendering layer}, we perform a novel optimization in latent space to regularize our reconstructed normals to be consistent with the manifold of natural shapes. To reconstruct the full 3D shape, we use PointNet++~\cite{qi2017pointnet++} with novel mechanisms to map normal features to a consistent 3D space, new loss functions for training and architectural changes that exploit surface normals for better recovery of 3D shape.

Since acquisition of transparent shapes is a laborious process, it is extremely difficult to obtain large-scale training data with ground truth \cite{jonathanacquisition}. Thus, we render a synthetic dataset, using a custom GPU-accelerated ray tracer. To avoid category-specific priors, we render images of random shapes under a wide variety of natural environment maps. 
%We scan ground truth 3D shapes of transparent objects to quantitatively and qualitatively evaluate our method, with extensive ablations of our choices. 
On both synthetic and real data, the benefits of our physically-based network design are clearly observed. Indeed, we posit that such physical modeling eases the learning for a challenging problem and improves generalization to real images. Figures \ref{fig:teaser} and \ref{fig:renderedKitten} show example outputs on real and synthetic data. All code and data will be publicly released.

To summarize, we propose the following contributions that solve the problem of transparent shape reconstruction with a limited number of unconstrained views:
\begin{tight_itemize}
\vspace{-0.2cm}
\item A physically-based network for surface normal reconstruction with a novel differentiable rendering layer and cost volume that imbibe insights from image formation.
\item A physically-based 3D point cloud reconstruction that leverages the above surface normals and rendering layer.
\item Strong experimental demonstration using a photorealistically rendered large-scale dataset for training and a small number of mobile phone photographs for evaluation.
\end{tight_itemize}

%Further, we posit that such physical modeling improves generalization, which is crucial in our case since it is extremely difficult to obtain ground truth shape for transparent objects [6] and simulations are the only scalable training resource. The challenge of designing network modules for complex light paths also affords unique opportunities, for example, we suggest reconstruction might benefit from observable signal arising from surfaces not visible in the camera.

\vspace{-0.2cm}
\section{Related Work}

\vspace{-0.2cm}
\paragraph{Multiview stereo}
%Since we use multiple images to obtain shape, multiview stereo (MVS) is related, where traditional approaches \cite{Seitz2006} as well as deep networks \cite{Yao2019} have achieved impressive results. 
Traditional approaches \cite{Seitz2006} and deep networks \cite{Yao2019} for multiview stereo have achieved impressive results. A full review is out of our scope, but we note that they assume photoconsistency for opaque objects and cannot handle complex light paths of transparent shapes. 
%without significant extensions, as proposed in this paper.

\vspace{-0.4cm}
\paragraph{Theoretical studies}
In seminal work, Kutulakos and Steger \cite{lightpath} characterize the extent to which shape may be recovered given the number of bounces in refractive (and specular) light paths.
% determining that three views suffice for two-bounce paths and enough views can reduce ambiguities to a discrete set for longer paths.
Chari and Sturm \cite{Chari2013} further constrain the system of equations using radiometric cues.
%, also studying the two-bounce case in particular. 
Other works study motion cues \cite{BenEzra2003,Morris2005} or parametric priors \cite{Tsai2015}.
%Some works consider motion cues \cite{BenEzra2003,Morris2005}, while Tsai et al.~\cite{Tsai2015} study the depth-normal ambiguity from a single light ray and propose parametric priors to resolve it. 
We derive inspiration from such works to incorporate physical properties of image formation, by accounting for refractions, reflections and total internal reflections in our network design.
%which acts as a powerful data-driven prior to disambiguate complex light paths.

\vspace{-0.4cm}
\paragraph{Controlled acquisition}
%Wetzstein et al.~use light field probes that encode both spatial and angular variations as reference patterns to recover normals and sparse 3D points for thin refractive objects \cite{Wetzstein2011}.
Special setups have been used in prior work, such as light field probes \cite{Wetzstein2011}, polarimetry \cite{Cui2017,Huynh2010,Miyazaki2005}, transmission imaging \cite{Kim2017}, scatter-trace photography \cite{Morris2007}, time-of-flight imaging \cite{Tanaka2016} or tomography \cite{Trifonov2006}. 
An external liquid medium \cite{Han2018} or moving spotlights in video \cite{Yeung2011} have been used too.
%Han et al.~\cite{Han2018} alleviate complex refractions by modifying the incident light path using an external liquid medium. A video sequence with a moving spotlight and a mirror sphere is used to reconstruct transparent shapes by Yeung et al.~\cite{Yeung2011}.
Wu et al.~\cite{wu2018full} also start from a visual hull like us, to estimate normals and depths from multiple views acquired using a turntable-based setup with two cameras that image projected stripe patterns in a controlled environment. A projector-camera setup is also used by \cite{Qian2016}. 
%along with position-normal consistency.
In contrast to all of the above works, we only require unconstrained natural images, even obtainable with a mobile phone camera, to reconstruct transparent shapes.

\vspace{-0.4cm}
\paragraph{Environment matting}
%Environment matting captures a map of refraction and reflection properties of an object that allows it to be efficiently composited in new scenes \cite{Zongker1999,Chuang2000}. It requires controlled acquisition through multiple projector screens that display a large number of predefined patterns. 
Environment matting uses a projector-camera setup to capture a composable map \cite{Zongker1999,Chuang2000}. Subsequent works have extended to mutliple cameras \cite{Matusik2002}, natural images \cite{Wexler2002}, frequency \cite{Zhu2004} or wavelet domains \cite{Peers2003}, with user-assistance \cite{yeungetal}, compressive sensing to reduce the number of images \cite{Duan2015,Qian2015} or deep network to predict the refractive flow from a single image \cite{chen2019learning}. In contrast, we use a small number of unconstrained images acquired with a mobile phone in arbitrary scenes, to produce full 3D shape.

\vspace{-0.4cm}
\paragraph{Reconstruction from natural images}
Stets et al.~\cite{stets2019single} propose a black-box network to reconstruct depth and normals from a single image.
%but do not recover a full 3D shape of high quality or explicitly account for transparent image formation. 
Shan et al.~\cite{Shan2012} recover height fields in controlled settings, while Yeung et al.~\cite{Yeung2014} have user inputs to recover normals. In contrast, we recover high-quality full 3D shapes and normals using only a few images of transparent objects, by modeling the physical basis of image formation in a deep network.

\vspace{-0.4cm}
\paragraph{Refractive materials besides glass}
%A differentiable renderer that can handle translucent materials is proposed by Che etal.~\cite{Che2018}, while neural volumes have also been proposed recently \cite{neuralvolumes}. 
Polarization \cite{Chen2007}, differentiable rendering \cite{Che2018} and neural volumes \cite{neuralvolumes} have been used for translucent objects, while specular objects have been considered under similar frameworks as transparent ones \cite{Ihrke2010,Zuo2015}. Gas flows \cite{Atcheson2008,Ji2013}, flames \cite{Ihrke2004,Wu2015} and fluids \cite{Gregson2012,Qian2017,Zhang2014} have been recovered, often in controlled setups.
%Atcheson et al.~\cite{Atcheson2008,} consider temporal acquisition of shapes of gas flows, while Ji et al.~\cite{Ji2013} use light path approximations. Ihrke et al.~\cite{Ihrke2004} use tomography to reconstruct flames, while color temperature is used by Wu et al.~\cite{Wu2015}. Qian et al.~\cite{Qian2017} reconstruct dynamic fluid surfaces, Zhang et al.~\cite{Zhang2014} undistort geometry under such moving interfaces and Gregson et al.~\cite{Gregson2012} do so for mixing fluids. 
Our experiments are focused on glass, but similar ideas might be applicable for other refractive media too.

\section{Method}
\label{sec:method}
%\noindent\MC{Some of this eventually moves to introduction.}

\vspace{-0.2cm}
\paragraph{Setup and assumptions}
Our inputs are $V$ images $\{I_{v}\}_{v=1}^V$ of a transparent object with known refractive index (IoR), along with segmentation masks $\{M_{v}\}_{v=1}^{V}$. We assume a known and distant, but otherwise arbitrary, environment map $E$. The output is a point cloud reconstruction $\mathcal{P}$ of the transparent shape. Note that our model is different from (3-2-2) triangulation \cite{Kutulakos2005} that requires two reference points on each ray for reconstruction, leading to a significant relaxation over prior works \cite{wu2018full,yeungetal} that need active lighting, carefully calibrated devices and controlled environments. We tackle this severely ill-posed problem through a novel physically-based network that models the image formation in transparent objects over three sub-tasks: shape initialization, cost volume for normal estimation and shape reconstruction.

\begin{figure}[t]
\centering
\includegraphics[width=3.3in]{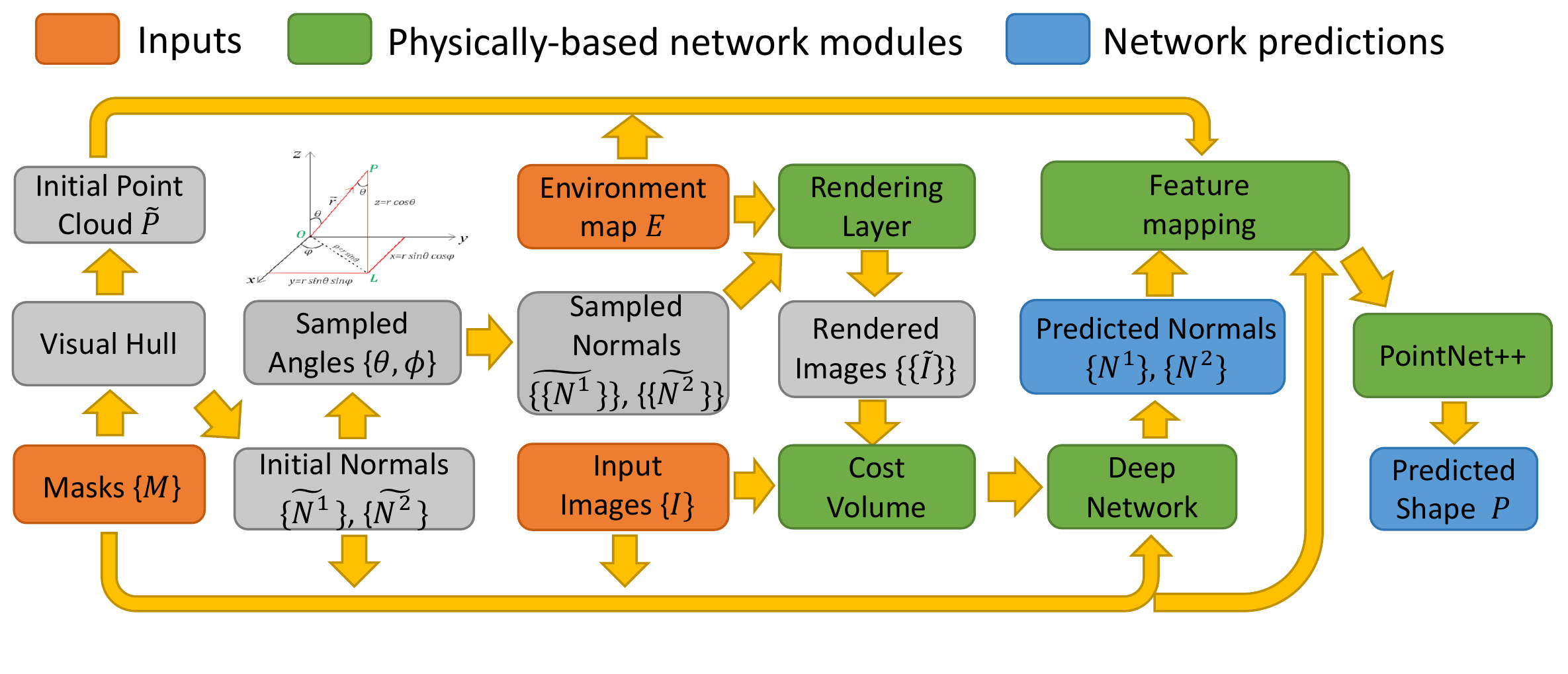}
\vspace{-0.7cm}
\caption{Our framework for transparent shape reconstruction.}
\label{fig:framework}
\vspace{-0.3cm}
\end{figure}

To simplify the problem and due to GPU memory limits, we consider light paths with only up to two bounces, that is, either the light ray gets reflected by the object once before hitting the environment map or it gets refracted by it twice before hitting the environment map. This is not a severe limitation -- more complex regions stemming from total internal reflection or light paths with more than two bounces are masked out in one view, but potentially estimated in other views. The overall framework is summarized in Figure \ref{fig:framework}.

\vspace{-0.4cm}
\paragraph{Shape initialization}
%\label{subsec:shapeInit}
We initialize the transparent shape with a visual hull  \cite{Kutulakos2000}. 
%Multi-view stereo methods based on photo-consistency loss can only handle opaque objects, usually with Lambertian material. While numerous learning or optimization-based methods reconstruct depth maps and surface normals from a single image, susceptibility to scale ambiguity makes them unsuited to multi-view reconstruction. 
%%On the other hand, the visual hull provides a constraint to prevent the reconstructed shape from drifting away from the ground-truth location. 
While a visual hull method cannot reconstruct some concave or self-occluded regions, it suffices as initialization for our network. We build a 3D volume of size $128^3$ and project  segmentation masks from $V$ views to it. Then we use marching cubes to reconstruct the hull and loop L3 subdivision to obtain smooth surfaces. 

\subsection{Normal Reconstruction}  
% \begin{figure*}
% \centering
% \includegraphics[width=6in]{Figures/normalNetwork.pdf}
% \caption{The network architecture for normal reconstruction. The blue color represents $\mathbf{NNet}$ and green color represents $\mathbf{FNet}$. $IX_1$-$OX_2$-$WX_3$-$SX_4$ represents a convolutional layer with input channel $X_1$, output channel $X_2$, kernel size $X_3$ and stride $X_4$. $UX_{5}$ represents bilinear upsampling layer with scale factor $X_{5}$ }
% \vspace{-0.2cm}
% \label{fig:normalNetwork}
% \end{figure*}
\begin{figure*}[!!t]
\begin{center}
\begin{minipage}[t]{0.65\linewidth}
\raisebox{-4cm}
{\includegraphics[width=0.98\linewidth]{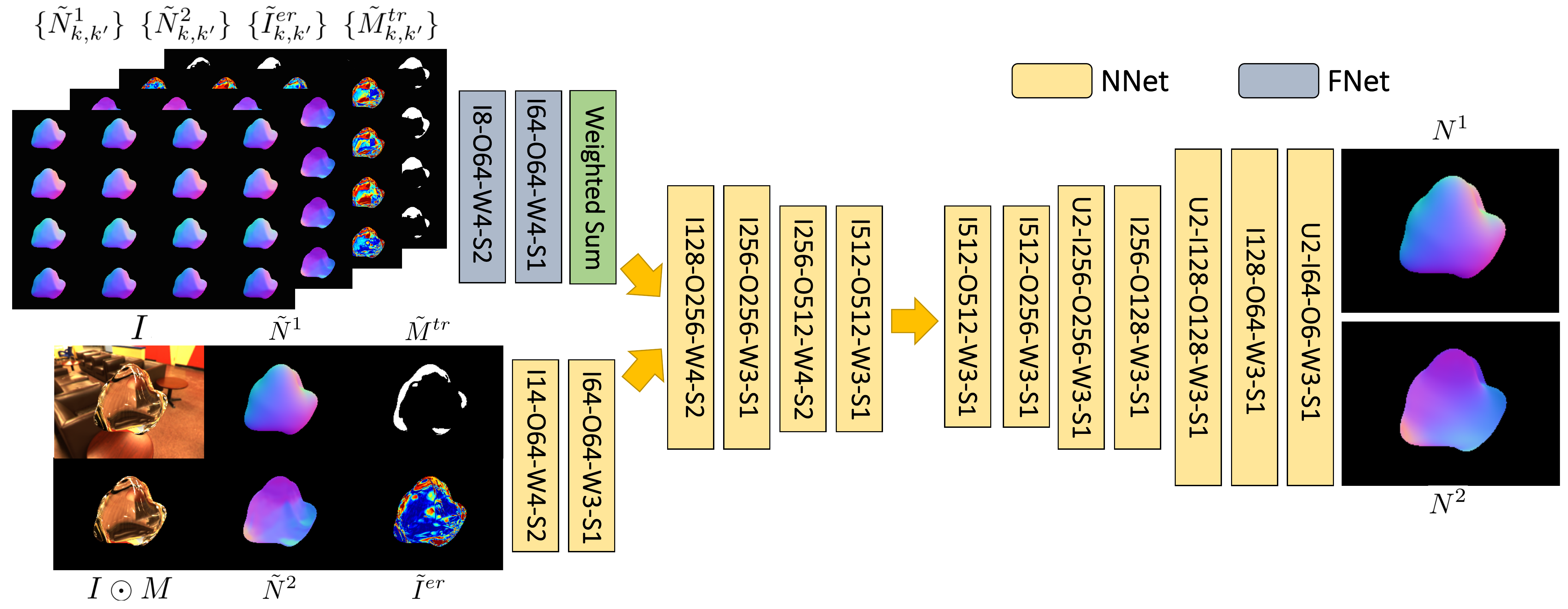}}
%{\includegraphics[width=0.98\linewidth]{Figures/NormalRecon.pdf}}
\end{minipage}\hfill
\begin{minipage}[t]{0.33\linewidth}
\caption{\small
The network architecture for normal reconstruction. Yellow blocks represent $\mathbf{NNet}$ and blue blocks represent $\mathbf{FNet}$. $IX_1$-$OX_2$-$WX_3$-$SX_4$ represents a convolutional layer with input channel $X_1$, output channel $X_2$, kernel size $X_3$ and stride $X_4$. $UX_{5}$ represents bilinear upsampling layer with scale factor $X_{5}$. 
\label{fig:normalNetwork}
}
\end{minipage}
\end{center}
\vspace{-0.8cm}
\end{figure*}

A visual hull reconstruction from limited views might be inaccurate, besides missed concavities. We propose to reconstruct high quality normals by estimating correspondences between the input image and the environment map. This is a very difficult problem, since different configurations of transparent shapes may lead to the same appearance. Moreover, small perturbations of normal directions can cause pixel intensities to be completely different. Thus, strong shape priors are necessary for a high quality reconstruction, which we propose to learn with a physically-inspired deep network.

\begin{figure}
\centering
\includegraphics[width=0.45\textwidth]{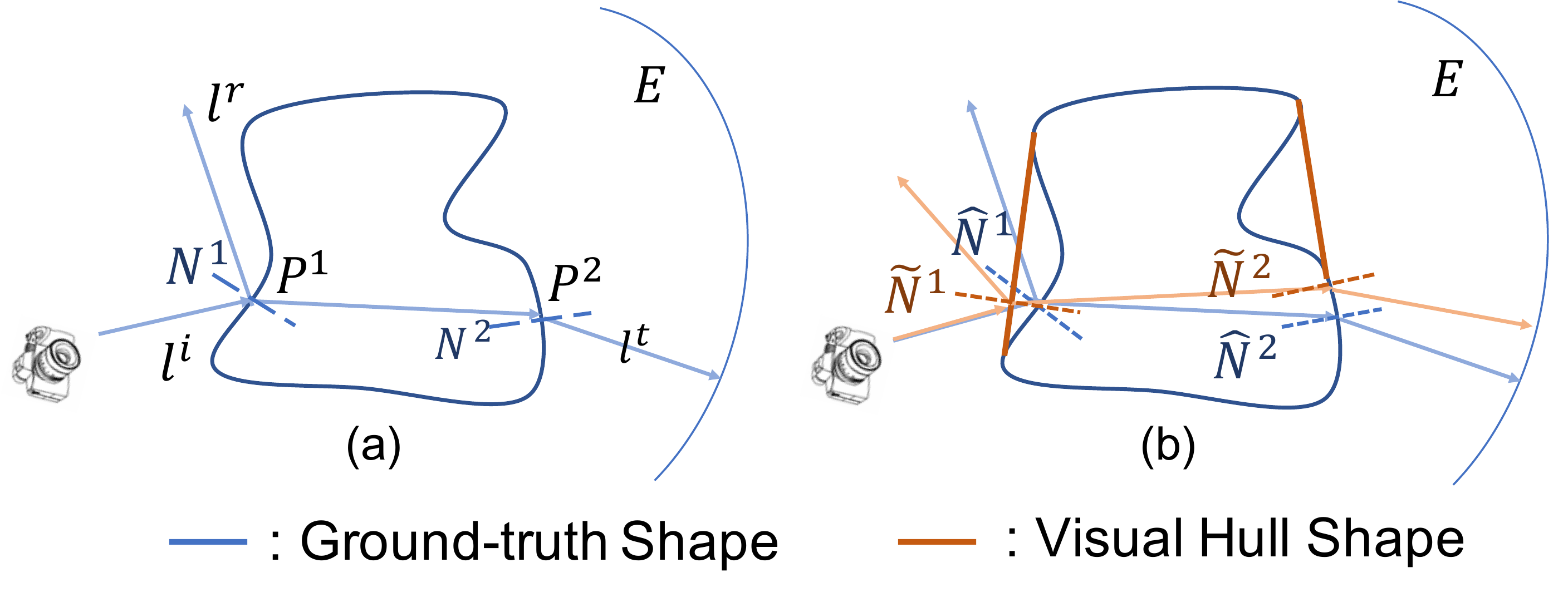}
\caption{(a) Illustration of the first and second normal ($N^1$ and $N^2$), the first and second hit points ($P^1$ and $P^2$), and the reflection and refraction modeled by our deep network. (b) Illustration of visual hull ($\tilde{N}^1$, $\tilde{N}^{2}$) and ground-truth normals ($\hat{N}^1$, $\hat{N}^{2}$).}
\vspace{-0.2cm}
\label{fig:normalDef}
\end{figure}

\vspace{-0.4cm}
\paragraph{Basic network} 
Our basic network architecture for normal estimation is shown in Figure \ref{fig:normalNetwork}. The basic network structure consists of one encoder and one decoder. The outputs of our network are two normal maps $N^1$ and $N^2$, which are the normals at the first and second hit points $P^1$ and $P^2$ for a ray backprojected from camera passing through the transparent shape, as illustrated in Figure \ref{fig:normalDef}(a). The benefit of modeling the estimation through $N^{1}$ and $N^{2}$ is that we can easily use a network to represent complex light transport effects without resorting to ray-tracing, which is time-consuming and difficult to treat differentiably. In other words, given $N^{1}$ and $N^{2}$, we can directly compute outgoing ray directions after passage through the transparent object. The inputs to our network are the image $I$, the image with background masked out $I\odot M$ and the $\tilde{N}^{1}$ and $\tilde{N}^{2}$ of the visual hull (computed off-line by ray tracing). We also compute $\hat{N}^{1}$ and $\hat{N}^{2}$ of the ground-truth shape for supervision. The definition of $\tilde{N}^1$, $\tilde{N}^{2}$ and $\hat{N}^1$, $\hat{N}^{2}$ are visualized in Figure \ref{fig:normalDef}(b). The basic network estimates:
\begin{equation}
\small
N^{1}, N^{2} = \mathbf{NNet}(I, I\odot M, \tilde{N}^{1}, \tilde{N}^{2} )
\end{equation}
The loss function is simply the $L_2$ loss for $N^1$ and $N^2$. 
\begin{equation}
\small
\mathcal{L}_{N} = ||N^{1} - \hat{N}^{1}||_{2}^2 +  ||N^{2} - \hat{N}^{2}||_{2}^2
\end{equation}

\vspace{-0.4cm}
\paragraph{Rendering layer} 
Given the environment map $E$, we can easily compute the incoming radiance through direction $l$ using bilinear sampling. This allows us to build a differentiable rendering layer to model the image formation process of refraction and reflection through simple local computation. As illustrated in Figure~\ref{fig:normalDef}(a), for every pixel in the image, the incident ray direction $l^{i}$ through that pixel can be obtained by camera calibration. The reflected and refracted rays $l^{r}$ and $l^{t}$ can be computed using $N^{1}$ and $N^{2}$, following Snell's law. \ZL{Our rendering layer implements the full physics of an intersection, including the intensity changes caused by the Fresnel term $\mathcal{F}$ of the refractive material. More formally, with some abuse of notation, let $L^{i}$, $L^{r}$ and $L^{t}$ be the radiance of incoming, reflected and refracted rays. We have}
\begin{eqnarray}
\small
\mathcal{F} = \frac{1}{2}\left(\frac{l^i \cdot N - \eta l^t\cdot N}{l^i\cdot N + \eta l^t\cdot N} \right)^2 \!\!\!\!\!\!\!\!&+&\!\!\!\!\!\! \frac{1}{2}\left(\frac{\eta l^i\cdot N - l^t\cdot N}{\eta l^i\cdot N + l^t\cdot N}\right)^2. \nonumber \\
\small
L^{r} = \mathcal{F}\cdot L^{i}, &&\!\!\!\!\!\!\!\!  L^{t} = (1 - \mathcal{F})\cdot L^{i} \nonumber
\end{eqnarray}

Due to total internal reflection, some rays entering the object may not be able to hit the environment map after one more bounce, for which our rendering layer returns a binary mask, $M^{tr}$. With $I^r$ and $I^t$ representing radiance along the directions $l^r$ and $l^t$, the rendering layer models the image formation process for transparent shapes through reflection, refraction and total internal reflection:
\begin{equation}
\small
I^{r}, I^{t}, M^{tr} = \mathbf{RenderLayer}(E, N^1, N^2).
\label{eq:renderLayer}
\end{equation}
%We neglect the contribution to pixel intensity from light path longer than 3. For regions without total reflection, we find such an approximation can closely match the ground-truth intensities. 
Our in-network rendering layer is differentiable and end-to-end trainable.
%can be built into the network for end-to-end training. 
But instead of just using the rendering loss as an extra supervision, we compute an error map based on rendering with the visual hull normals:
%$\tilde{I}^{er}$ by using our rendering layer to render the image from the visual hull normals $\tilde{N}^{1}$ and $\tilde{N}^{2}$ and comparing with the input image $I$:
\small
\begin{eqnarray}
\tilde{I}^{r}, \tilde{I}^{t}, \tilde{M}^{tr} &=& \mathbf{RenderLayer}(E, \tilde{N}^{1}, \tilde{N}^{2}) , 
\label{eq:errorMap1} \\
\tilde{I}^{er} &=& | I - (\tilde{I}^{r} + \tilde{I}^{t} ) | \odot M .
\label{eq:errorMap2}
\end{eqnarray}
\normalsize
This error map is used as an additional input to our normal reconstruction network, to help it better learn regions where the visual hull normals $\tilde{N}^{1}$ and $\tilde{N}^{2}$ may not be accurate: 
%We stack the $\tilde{I}^{er}$ and the total reflection mask $\tilde{M}^{tr}$ before sending them to the normal prediction network. 
\begin{equation}
\small
N^{1}, N^{2} = \mathbf{NNet}(I, I\odot M, \tilde{N}^{1}, \tilde{N}^{2}, \tilde{I}^{er}, \tilde{M}^{tr} )
\end{equation}
%With the predictions $N^1$ and $N^2$, the error image $I^{er}$ and total reflection mask $M^{tr}$ can be computed in a similar way. 
%\begin{eqnarray}
%I^{r}, I^{t}, M^{tr} &=& \mathbf{Renderer}(E, N^{1}, N^{2} ) \\
%I^{er} &=& ||I - (I^{r} + I^{t} )||^{2}_{2} \odot M
%\end{eqnarray}

% \begin{figure}
% \centering
% \includegraphics[width=2.5in]{Figures/normalDef_2.pdf}
% \caption{A demonstration of the first and second normal ($N^1$ and $N^2$), the first and second hit points ($P^1$ and $P^2$) ,and the reflection and refraction modeled by our deep network.}
% \vspace{-0.2cm}
% \label{fig:normalDef}
% \end{figure}

% \begin{figure}
% \centering
% \includegraphics[width=3.0in]{Figures/normalVHGt_2.pdf}
% \caption{A demonstration of visual hull normals $\tilde{N}^1$, $\tilde{N}^{2}$ and ground-truth normals $\hat{N}^1$, $\hat{N}^{2}$.}
% \label{fig:normalDef_vh_gt}
% \end{figure}

% \begin{figure*}
% \centering
% \includegraphics[width=6in]{Figures/pointNetwork.pdf}
% \caption{Our pipeline for point cloud reconstruction. $AX_1$-$SX_2$-$L(X_3, X_4)$-$RX_5$ represents a set abstraction layer with $X_1$ anchor points, $X_2$ sampled points, 2 fully connected layers with feature channel number $X_3$ and $X_4$ and sampling radius $0.5$. $IY_1$-$L(Y_2, Y_3)$ represents a unit PointNet with input channel number $Y_1$ and 2 fully connected layers with feature channel number $Y_2$ and $Y_3$. }
% \label{fig:pointNetwork}
% \end{figure*}

\begin{figure}
\centering
\includegraphics[width=3in]{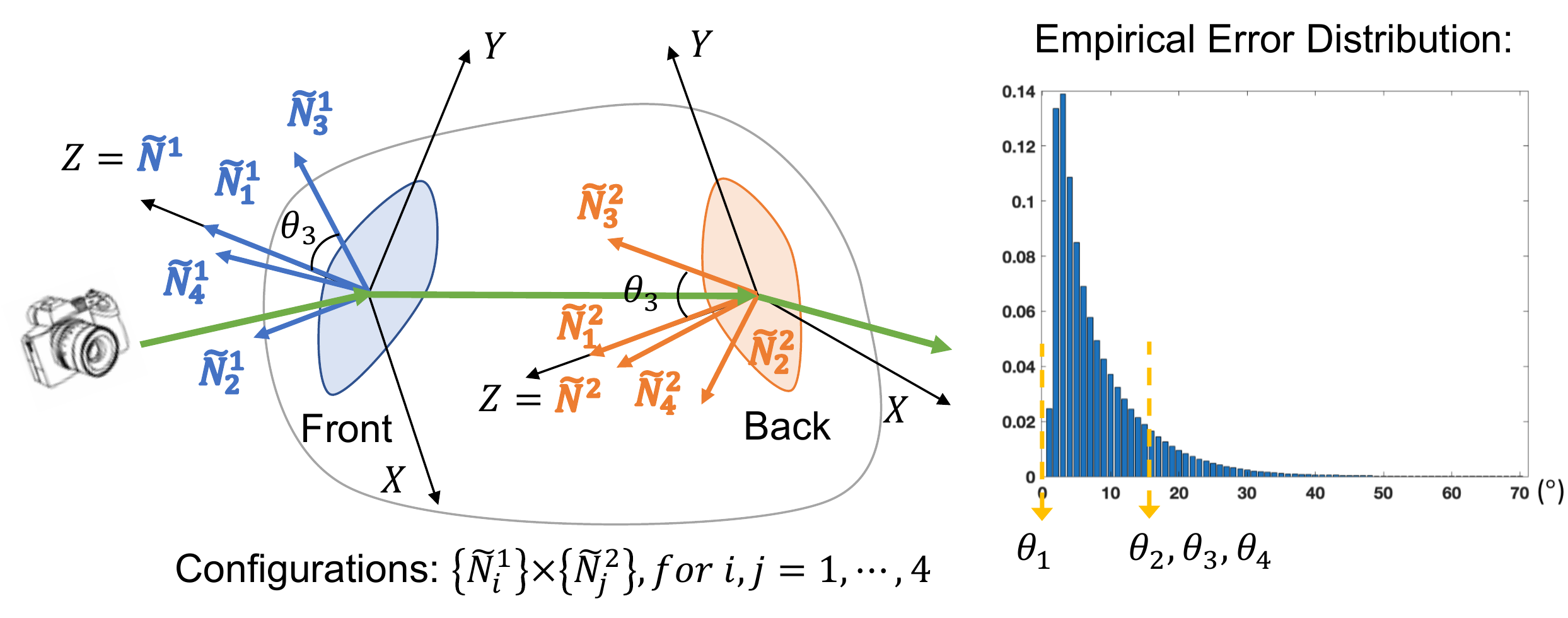}
\caption{We build an efficient cost volume by sampling directions around visual hull normals according to their error distributions. }
\vspace{-0.2cm}
\label{fig:costVolume}
\end{figure}

\begin{figure*}[!!t]
\begin{center}
\begin{minipage}[t]{0.60\linewidth}
\raisebox{-2.6cm}
{\includegraphics[width=0.98\linewidth]{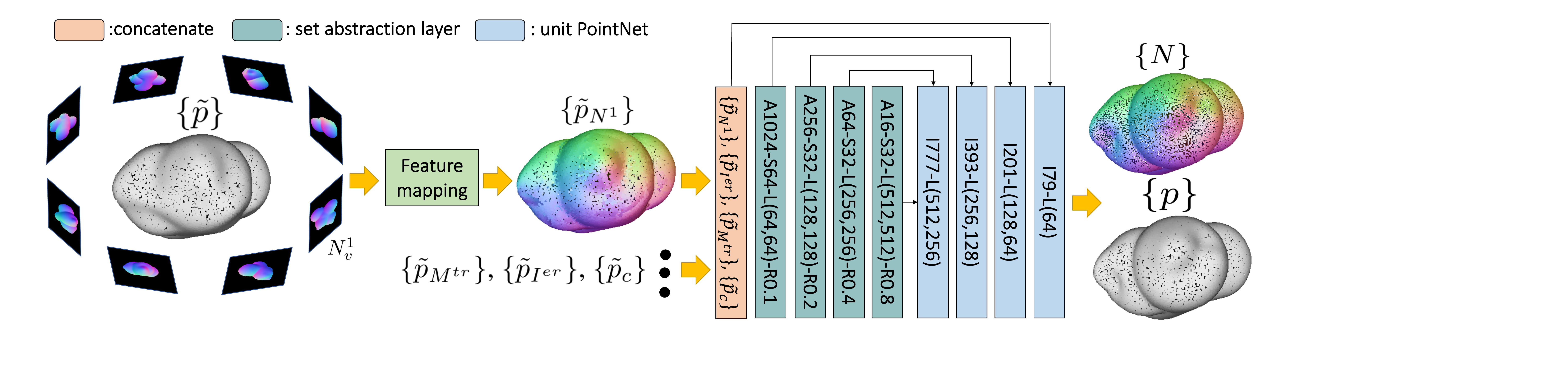}}
%{\includegraphics[width=0.98\linewidth]{Figures/NormalRecon.pdf}}
\end{minipage}\hfill
\begin{minipage}[t]{0.38\linewidth}
\caption{\small
Our method for point cloud reconstruction. $AX_1$-$SX_2$-$L(X_3, X_4)$-$RX_5$ represents a set abstraction layer with $X_1$ anchor points, $X_2$ sampled points, 2 fully connected layers with $X_3$, $X_4$ feature channels and sampling radius $X_5$. $IY_1$-$L(Y_2, Y_3)$ represents a unit PointNet with $Y_1$ input channels and 2 fully connected layers with $Y_2$, $Y_3$ feature channels.  
\label{fig:pointNetwork}
}
\end{minipage}
\end{center}
\vspace{-0.8cm}
\end{figure*}

\vspace{-0.4cm}
\paragraph{Cost volume} 
We now propose a cost volume to leverage the correspondence between the environment map and the input image. While cost volumes in deep networks have led to great success for multiview depth reconstruction of opaque objects, extension to normal reconstruction for transparent objects is non-trivial. The brute-force approach would be to uniformly sample the 4-dimensional hemisphere of $N^{1} \times N^{2}$, then compute the error map for each sampled normal. However, this will lead to much higher GPU memory consumption compared to depth reconstruction due to higher dimensionality of the sampled space. To limit memory consumption, we sample $N^{1}$ and $N^{2}$ in smaller regions around the initial visual hull normals $\tilde{N}^{1}$ and $\tilde{N}^{2}$, as shown in Figure \ref{fig:costVolume}. 
Formally, let $U$ be the up vector in bottom-to-top direction of the image plane. We first build a local coordinate system with respect to $\tilde{N}^{1}$ and $\tilde{N}^{2}$: 
\begin{eqnarray}
\small
Z = \tilde{N}^{i}, \;
Y = U - (U^{T}\cdot \tilde{N}^{i})  \tilde{N}^{i}, \;
X = \mathtt{cross}(Y, Z),
\end{eqnarray}
%\begin{eqnarray}
%\footnotesize
%Z &=& \tilde{N}^{i}, \\
%Y &=& \mathtt{Normalize}(U - (U^{T}\cdot \tilde{N}^{i})  \tilde{N}^{i}), \\
%X &=& \mathtt{cross}(Y, Z),
%\end{eqnarray}
where $Y$ is normalized and $i = 1, 2$. Let $\{\theta_{k}\}_{k=1}^{K}$, $\{\phi_{k}\}_{k=1}^{K}$ be the sampled angles. Then, the sampled normals are:
\begin{equation}
\tilde{N}^{i}_{k} = X \cos \phi_{k} \sin \theta_{k} + Y \sin \phi_{k} \sin \theta_{k} +  Z \cos \theta_{k}.
\end{equation}
\ZL{We sample the angles $\{\theta_{k}\}_{k=1}^{K}$, $\{\phi_{k}\}_{k=1}^{K}$ according to the error distribution of visual hull normals. The angles and distributions are shown in the supplementary material.} Since we reconstruct $N^{1}$ and $N^{2}$ simultaneously, the total number of configurations of sampled normals is $K\times K$. Directly using the $K^2$ sampled normals to build a cost volume is too expensive, so we use a learnable pooling layer to aggregate the features from each sampled normal configuration in an early stage. For each pair of $\tilde{N}^{1}_{k}$ and $\tilde{N}^{2}_{k'}$, we compute their total reflection mask $\tilde{M}^{tr}_{k, k'}$ and error map $\tilde{I}^{er}_{k, k'}$ using \eqref{eq:errorMap1} and \eqref{eq:errorMap2}, then perform a feature extraction:
%We stack them together and send to a network module for feature extraction: 
\begin{equation}
\small
F(k, k') = \mathbf{FNet}(\tilde{N}^{1}_{k}, \tilde{N}^{2}_{k'}, \tilde{I}^{er}_{k, k'}, \tilde{M}^{tr}_{k, k'}).
\end{equation}
We then compute the weighted sum of feature vectors $F(k, k')$ and concatenate them with the feature extracted from the encoder of $\mathbf{NNet}$ for normal reconstruction: 
\begin{equation}
\small
F = \sum_{k}^{K}\sum_{k'}^{K}\omega(k, k') F(k, k'),
\end{equation}
where $\omega(k, k')$ are positive coefficients with sum equal to 1, that are also learned during the training process. The detailed network structure is shown in Figure \ref{fig:normalNetwork}.

\vspace{-0.4cm}
\paragraph{Post processing} The network above already yields reasonable normal reconstruction. It can be further improved by optimizing the latent vector from the encoder to minimize the rendering error using the predicted normal $N^1$ and $N^2$:
%We define the loss function  $\mathcal{L}_{N}^{\text{Opt}}$ for the optimization process to be the $L_2$ error of the rendered image using the predicted normal $N^1$ and $N^2$
\begin{equation}
\small
%I^{r}, I^{t}, M^{tr} &=& \textbf{Renderer}(E, N^1, N^2) \\
\mathcal{L}_{N}^{\text{Opt}} = || (I - (I^r + I^t))\odot M^{tr} ||_{2}^{2},
\end{equation}
where $I_t, I_t, M^{tr}$ are obtained from the rendering layer \eqref{eq:renderLayer}. For this optimization, we keep the network parameters unchanged and only update the latent vector. 
%to minimize the differences between the input images and the image rendered with predicted normal $N^{1}$ and $N^{2}$ using our \ZL{differentiable rendering layer}. 
%While it can greatly reduce rendering error and noticeably improves normal reconstruction, such improvements cannot be achieved by 
Note that directly optimizing the predicted normal $N^{1}$ and $N^{2}$ without the deep network does not yield comparable improvements. This is due to our decoder acting as a regularization that prevents the reconstructed normal from deviating from the manifold of natural shapes during the optimization. Similar ideas have been used for BRDF reconstruction \cite{gao2019deep}.  %More details and comparisons are included in Section \ref{sec:experiments}. 

\subsection{Point Cloud Reconstruction} 
We now reconstruct the transparent shape based on the predictions of $\mathbf{NNet}$, that is, the normals, total reflection mask and rendering error. Our idea is to map the predictions from different views to the visual hull geometry. These predictions are used as input features for a point cloud reconstruction to obtain a full 3D shape. Our point cloud reconstruction pipeline is illustrated in Figure \ref{fig:pointNetwork}.

\vspace{-0.4cm}
\paragraph{Feature mapping}
We propose three options to map predictions from different views to the visual hull geometry. Let $\{\tilde{p}\}$ be the point cloud uniformly sampled from visual hull surfaces and $\mathcal{S}_{v}(\tilde{p}, h)$ be a function that projects the 3D point $\tilde{p}$ to the 2D image plane of view $v$ and then fetches the value of a function $h$ defined on image coordinates using bilinear sampling. Let $\mathcal{V}_{v}(\tilde{p})$ be a binary function that verifies if point $\tilde{p}$ can be observed from view $v$ 
%This is easy since have rendered the depth maps of visual hull from different views. 
and $\mathcal{T}_{v}(\tilde{p})$ be a transformation that maps a 3D point or normal direction in view $v$ to world coordinates. Let $\mathcal{C}_{v}(\tilde{p})$ be the cosine of the angle between the ray passing through $\tilde{p}$ and camera center. 

The first option is a feature $f$ that averages observations from different views. For every view $v$ that can see the point $\tilde{p}$, we project its features to the point and compute a mean: 
\begin{eqnarray}
\small
\tilde{p}_{N^{1}} = \frac{\scriptstyle\sum_{v}\mathcal{T}_{v}(\mathcal{S}_v(\tilde{p}, N_{v}^1) )\mathcal{V}_v(\tilde{p})}{\scriptstyle\sum_{v}\mathcal{V}_v(\tilde{p})}, \!\!\!&&\!\!\!
\tilde{p}_{I^{er}} = \frac{\scriptstyle\sum_{v}\mathcal{S}_v(\tilde{p}, I_{v}^{er})\mathcal{V}_{v}(\tilde{p})}{\scriptstyle\sum_{v}\mathcal{V}_v(\tilde{p})}, \nonumber\\
\tilde{p}_{M^{tr}} = \frac{\scriptstyle\sum_{v}\mathcal{S}_v(\tilde{p}, M_{v}^{tr})\mathcal{V}_{v}(\tilde{p})}{\scriptstyle\sum_{v}\mathcal{V}_v(\tilde{p})}, \!\!\!&&\!\!\!
\tilde{p}_{c} = \frac{\scriptstyle\sum_{v}\mathcal{C}_{v}(\tilde{p})\mathcal{V}_{v}(\tilde{p})}{\scriptstyle\sum_{v}\mathcal{V}_v(\tilde{p})}. \nonumber
\end{eqnarray}
We concatenate to get: $f = [\tilde{p}_{{N}^1}, \tilde{p}_{I^{er}}, \tilde{p}_{M^{tr}}, \tilde{p}_{c}]$. 

Another option is to select a view $v^{*}$ with potentially the most accurate predictions and compute $f$ using the features from only that view. We consider two view-selection strategies. The first is nearest view selection, in which we simply select $v^{*}$ with the largest $\mathcal{C}_{v}(\tilde{p})$. %The other is the rendering error based view selection, 
The other is to choose the view with the lowest rendering error and no total reflection, with the algorithm detailed in supplementary material. Note that although we do not directly map $N^{2}$ to the visual hull geometry, it is necessary for computing the rendering error and thus, needed for our shape reconstruction.

\begin{figure}
\centering
\includegraphics[width=3.0in]{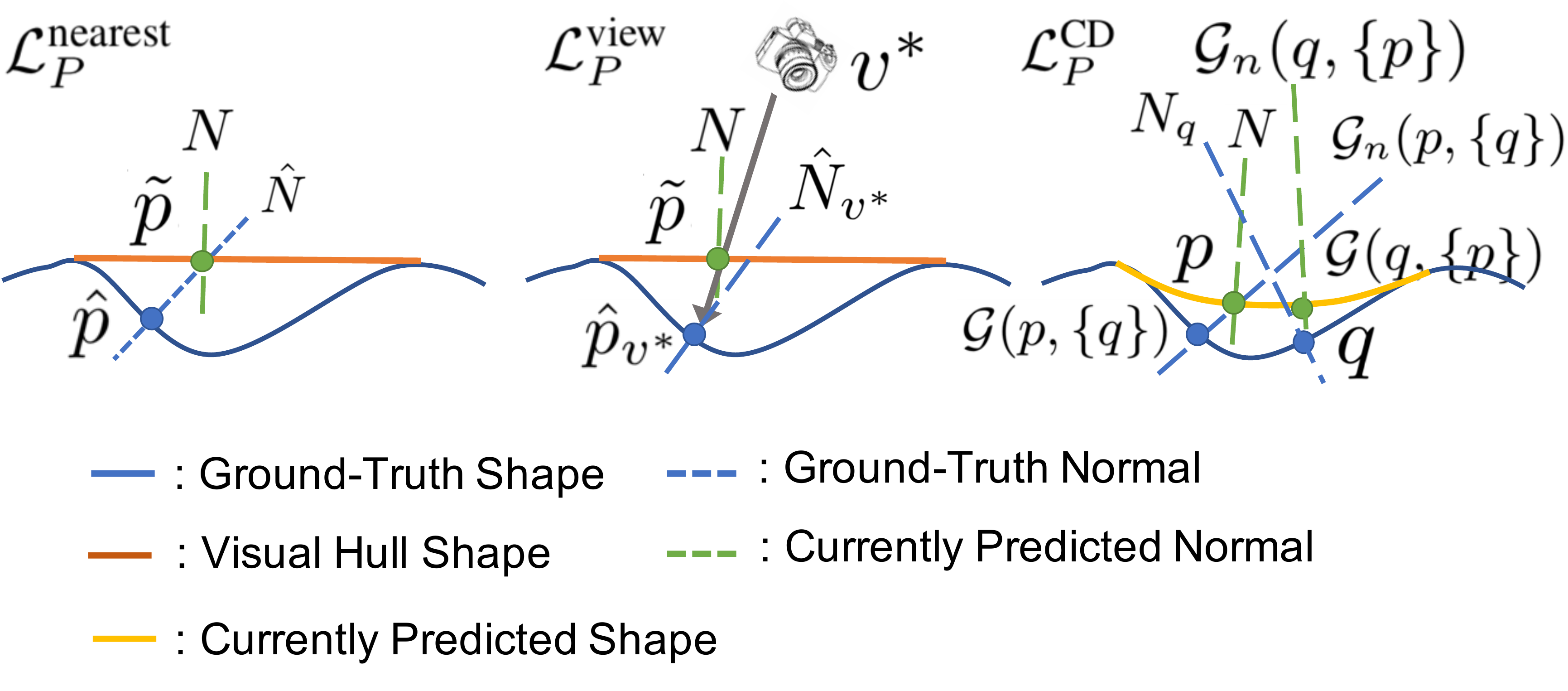}
\caption{A visualization of the loss functions for point cloud reconstruction. From left to the right are nearest $L_2$ loss $\mathcal{L}_{P}^{\text{nearest}}$, view-dependent $L_2$ loss $\mathcal{L}_{P}^{view}$ and chamfer distance loss $\mathcal{L}_{P}^{\text{CD} }$. }
\vspace{-0.2cm}
\label{fig:pointNetLossFuncs}
\end{figure}

\vspace{-0.4cm}
\paragraph{Point cloud refinement} 
We build a network following PointNet++ \cite{qi2017pointnet++} to reconstruct the point cloud of the transparent object. The input to the network is the visual hull point cloud $\{\tilde{p}\}$ and the feature vectors $\{ f \}$. The outputs are the normals \{$N$\} and the offset of visual hull points $\{\delta \tilde{p}\}$, with the final vertex position is computed as $p = \tilde{p} + \delta\tilde{p}$: 
\begin{equation}
\{\delta \tilde{p}\}, \{{N}\} = \mathbf{PNet}(\{\tilde{p}\}, \{ f \}).
\end{equation}
We tried three loss functions to train our PointNet++. The first loss function is the nearest $L_{2}$ loss $\mathcal{L}_{P}^{\text{nearest}}$. Let $\hat{p}$ be the nearest point to $\tilde{p}$ on the surface of ground-truth geometry and $\hat{N}$ be its normal. We compute the weighted sum of $L_2$ distance between our predictions ${p}$, ${N}$ and ground truth: 
\begin{equation}
\mathcal{L}_{P}^{\text{nearest} } = \sum_{\{{p}\}, \{ N \}}\lambda_{1}||{p} -  \hat{p}||_{2}^{2} + \lambda_{2}||{N} - \hat{N}||_{2}^{2}.
\end{equation}
The second loss function is a view-dependent $L_2$ loss $\mathcal{L}_{P}^{\text{view}}$. Instead of choosing the nearest point from ground-truth geometry for supervision, we choose the point from the best view $v^*$ by projecting its geometry into world coordinates:
\begin{equation}
\small
\hat{p}_{v^*}, \hat{N}_{v^*} = \left\{
\!\!\!
\renewcommand{\arraystretch}{1.2}
\begin{array}{cc}
{\scriptstyle \mathcal{T}_{v^{\!*}}(\mathcal{S}_{v^{\!*}}(\tilde{p}, \hat{P}^{1}_{v^{\!*}}) ),\mathcal{T}_{v^{\!*}}(\mathcal{S}_{v^{\!*}}(\tilde{p}, \hat{N}^{1}_{v^{\!*}}) )}, & v^* \neq 0 \\
\hat{p}, \hat{N},  & v^* = 0.
\end{array}
\right.\nonumber
\end{equation}
Then we have 
\begin{equation}
\mathcal{L}_{P}^{\text{view} } = \sum_{\{{p}\}, \{ N \}} \lambda_{1}||{p} - \hat{p}_{v^*}||_{2}^{2} + \lambda_{2}||{N} - \hat{N}_{v^*}||_{2}^{2}.
\end{equation}
The intuition is that since both the feature and ground-truth geometry are selected from the same view, the network can potentially learn their correlation more easily. The last loss function, $\mathcal{L}^{\text{CD}}_{P}$, is based on the Chamfer distance. Let $\{ q \}$ be the set of points uniformly sampled from the ground-truth geometry, with normals $N_q$. Let $\mathcal{G}(p, \{ q \})$ be a function which finds the nearest point of $p$ in the point set \{$q$\} and function $\mathcal{G}_{n}(p, \{ q \})$ return the normal of the nearest point. The Chamfer distance loss is defined as 
\vspace{-0.1cm}
\small
\begin{eqnarray}
\mathcal{L}^{\text{CD}}_{P}\!\!\!\!\!\!&=&\!\!\!\!\!\!\!\!
\sum_{\{{p}\}, \{ N \}}\!\!\!\frac{\lambda_{1}}{2}||{p}\!-\! \mathcal{G}({p}, \{q\})|| \!+\! \frac{\lambda_{2}}{2}||{N}\!-\!\mathcal{G}_{n}({p}, \{q\}) || + \nonumber \\ 
\!\!\!\!\!\!\!\!\!&&\!\!\!\!\!\!\!\!\!\!\!\!\sum_{\{q\}, \{ N_q \}}\!\! \frac{\lambda_{1}}{2}||q\!-\!\mathcal{G}(q, \{{p}\} )|| \!+\! \frac{\lambda_{2}}{2}||N_q \!-\!\mathcal{G}_{n}(q, \{{p}\}) ||.
\vspace{-0.1cm}
\label{eq:chamferLoss}
\end{eqnarray}
\normalsize
Figure \ref{fig:pointNetLossFuncs} is a demonstration of the three loss functions. In all our experiments, we set $\lambda_{1} = 200$ and $\lambda_{2} = 5$. 

%Our network consists of one encoder and one decoder, as 
Our network, shown in Figure \ref{fig:pointNetwork}, makes several improvements over standard PointNet++. First, we replace max-pooling with average-pooling to favor smooth results. Second, we concatenate normals \{${N}$\} to all skip connections to learn details. Third, we augment the input feature of set abstraction layer with the difference of normal directions between the current and center points. Section \ref{sec:experiments} and supplementary material show the impact of our design choices.

\section{Experiments}
\label{sec:experiments}
%In this section, we will present our dataset generation pipeline, the implementation details, and our 3D reconstruction results on both real and synthetic transparent objects.

\begin{comment}
\begin{table}[t]
\footnotesize
\renewcommand{\arraystretch}{1.1}
\centering
\begin{tabular}{|l|c|c|}
\hline
& $\{\theta_{k}\}_{k=1}^{4}$ & $\{\phi_{k}\}_{k=1}^{4}$ \\
\hline
5 views & $0^{\circ}$,$25^{\circ}$,$25^{\circ}$,$25^{\circ}$ & $0^{\circ}$,$0^{\circ}$,$120^{\circ}$,$240^{\circ}$  \\
 \hline
10 views & $0^{\circ}$,$15^{\circ}$,$15^{\circ}$,$15^{\circ}$  & $0^{\circ}$,$0^{\circ}$,$120^{\circ}$,$240^{\circ}$   \\
\hline
20 views & $0^{\circ}$,$10^{\circ}$,$10^{\circ}$,$10^{\circ}$ &$0^{\circ}$,$0^{\circ}$,$120^{\circ}$,$240^{\circ}$ \\
\hline
\end{tabular}
\vspace{0.1cm}
\caption{The sampled angles for building cost volume. We set the sampled angles according to the normal error of visual hull reconstructed by different number of views. }
\label{tab:sampledAngles} 
\vspace{-0.3cm}
\end{table}
\end{comment}

%\vspace{-0.1cm}
%\subsection{Training Data and Implementation Details} 
%\label{sec:dataset}

\vspace{-0.1cm}
\paragraph{Dataset}
We procedurally generate random scenes following \cite{li2018svbrdf,xu2018deep} rather than use shape repositories \cite{chang2015shapenet}, to let the model be category-independent. To remove inner structures caused by shape intersections and prevent false refractions, we render 75 depth maps and use PSR \cite{kazhdan2006poisson} to fuse them into a mesh, with L3 loop subdivision to smooth the surface. We implement a physically-based GPU renderer using NVIDIA OptiX \cite{OptiX}. With $1499$ HDR environment maps of \cite{gardner2017learning} for training and $424$ for testing, we render 3000 random scenes for training and 600 for testing. The IoR of all shapes is set to 1.4723, to match our real objects. Our experiments also include sensitivity analysis to characterize the behavior of the network when the test-time IoR differs from this value.

\vspace{-0.4cm}
\paragraph{Implementation Details} 
When building the cost volume for normal reconstruction, we set the number of sampled angles $K$ to be 4. Increasing the number of sampled angles will drastically increase the memory consumption and does not improve the normal accuracy. We sample $\phi$ uniformly from 0 to $2\pi$ and sample $\theta$ according to the visual hull normal error. The details are included in the supplementary material. We use Adam optimizer to train all our networks. The initial learning rate is set to be $10^{-4}$ and we halve the learning rate every 2 epochs. All networks are trained over 10 epochs.

%We first randomly sample 100 scenes from our synthetic dataset and compute the angles between visual hull normals and ground-truth normals. We set one $\theta$ value to be 0 and the other to larger than 85\% of angles between visual hull normal and ground-truth normals. Table \ref{tab:sampledAngles} summarizes the configurations of $\{\theta\}$ and $\{\phi\}$ angles for different number of views.

\begin{figure*}[t]
\centering
\includegraphics[width=6.5in]{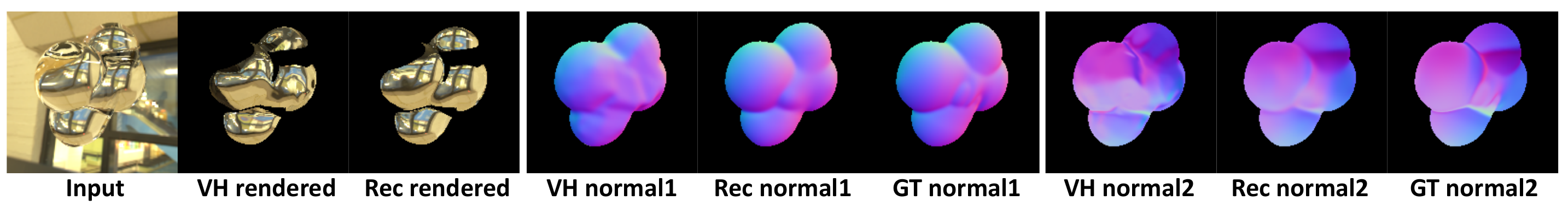}
\vspace{-0.1cm}
\caption{An example of 10 views normal reconstruction from our synthetic dataset. The region of total reflection has been masked out in the rendered images. }
\label{fig:normalQuality}
\vspace{-0.3cm}
\end{figure*}

\begin{figure*}
\centering
\includegraphics[width=6.5in]{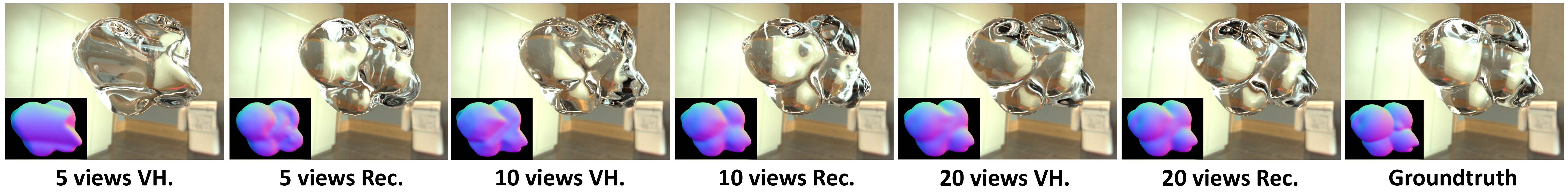}
\caption{Our transparent shape reconstruction results from 5 views, 10 views and 20 views from our synthetic dataset. The images rendered with our reconstructed shapes are much closer to the ground-truth compared with images rendered with the visual hull shapes. The inset normals are rendered from the reconstructed shapes.}
\label{fig:synDiffView}
\vspace{-0.3cm}
\end{figure*}

\subsection{Ablation Studies on Synthetic Data} 
\label{sec:abl_syn}

\begin{table}[t]
\footnotesize
\renewcommand{\arraystretch}{1.2}
\setlength{\tabcolsep}{4pt}
\centering
\begin{tabular}{l|c|c|c|c|c|c}
\hline
& \multirow{2}{*}{$\mathtt{vh10}$} & \multirow{2}{*}{$\mathtt{basic}$}  & \multirow{2}{*}{$\mathtt{wr}$} & \multirow{2}{*}{$\mathtt{wr}$+$\mathtt{cv}$} & $\mathtt{wr}$+$\mathtt{cv}$ & $\mathtt{wr}$+$\mathtt{cv}$ \\
    & & & & & +$\mathtt{op}$ & $\mathtt{var.}$ $\mathtt{IoR}$ \\
\hline
$N^1$ median ($^\circ$) &5.5 & 3.5 & 3.5 &3.4 & \textbf{3.4} & 3.6 \\
\hline
$N^{1}$ mean ($^\circ$) &7.5 & 4.9 & 5.0 &4.8 & \textbf{4.7} & 5.0 \\
\hline 
$N^2$ median ($^\circ$) & 9.2 & 6.9 & 6.8 & 6.6 & \textbf{6.2} & 7.3 \\
\hline 
$N^{2}$ mean ($^\circ$) & 11.6 & 8.8 & 8.7 & 8.4 & \textbf{8.1} & 9.1\\
\hline
Render Err.($10^{-2}$) & 6.0 & 4.7 & 4.6 & 4.4 & \textbf{2.9} & 5.5\\
\hline
\end{tabular}
\vspace{0.05cm}
\caption{Quantitative comparisons of normal estimation from 10 views. $\mathtt{vh10}$ represents the initial normals reconstructed from 10 views visual hull. $\mathtt{wr}$ and $\mathtt{basic}$  are our basic encoder-decoder network with and without rendering error map ($I^{er}$) and total reflection mask ($M^{tr}$) as inputs. $\mathtt{wr}$+$\mathtt{cv}$ represents our network with cost volume.  $\mathtt{wr}$+$\mathtt{cv}$+$\mathtt{op}$ represents the predictions after optimizing the latent vector to minimize the rendering error. $\mathtt{wr}$+$\mathtt{cv}$ $\mathtt{var.}$ $\mathtt{IoR}$ represents sensitivity analysis for IoR, explained in text. }
\vspace{-0.2cm}
\label{tab:normalQuan}
\end{table}

\vspace{-0.1cm}
\paragraph{Normal reconstruction} The quantitative comparisons of 10 views normal reconstruction are summarized in Table \ref{tab:normalQuan}. We report 5 metrics: the median and mean angles of the first and the second normals ($N^1$, $N^2$), and the mean rendering error ($I^{er}$). We first compare the normal reconstruction of the basic encoder-decoder structure with ($\mathtt{wr}$) and without rendering error and total reflection mask as input ($\mathtt{basic}$). While both networks greatly improve the normal accuracy compared to visual hull normals ($\mathtt{vh10}$), adding rendering error and total reflection mask as inputs can help achieve overall slightly better performances. Next we test the effectiveness of the cost volume ($\mathtt{wr}$+$\mathtt{cv}$). Quantitative numbers show that adding cost volume achieves better results, which coincides with our intuition that finding the correspondences between input image and the environment map can help our normal prediction. Finally we optimize the latent vector from the encoder by minimizing the rendering error ($\mathtt{wr}$+$\mathtt{cv}$+$\mathtt{op}$). It significantly reduces the rendering error and also improves the normal accuracy. Such improvements cannot be achieved by directly optimizing the normal predictions $N^1$ and $N^2$ in the pixel space. Figure \ref{fig:normalQuality} presents normal reconstruction results from our synthetic dataset. Our normal reconstruction pipeline obtains results of much higher quality compared with visual hull method. Ablation studies of 5 views and 20 views normal reconstruction and the optimization of latent vector are included in the supplementary material.

\begin{table}[t]
\footnotesize
\renewcommand{\arraystretch}{1.2}
\setlength{\tabcolsep}{1pt}
\centering
\begin{tabular}{l|c|c|c|c}
\hline
& CD($10^{-4}$) & CDN-mean($^\circ$) & CDN-med($^\circ$) & Metro($10^{-3}$) \\
\hline
\small{$\mathtt{vh10}$} & 5.14 & 7.19 & 4.90 & 15.2  \\
\hline
\small{$\mathtt{RE}$-$\mathcal{L}_{P}^{\text{nearest}}$} &2.17  &6.23 &4.50 &7.07 \\
%\hline 
\small{$\mathtt{RE}$-$\mathcal{L}_{P}^{\text{view}}$} &2.15 &6.51 &4.76  &6.79 \\
%\hline 
\small{$\mathtt{RE}$-$\mathcal{L}_{P}^{\text{CD}}$}& \textbf{2.00} & \textbf{6.02} & \textbf{4.38} & \textbf{5.98} \\ 
\hline
\hline 
\small{$\mathtt{NE}$-$\mathcal{L}_{P}^{\text{CD}}$} &2.04 &6.10 &4.46 & 6.02 \\ 
\hline
\small{$\mathtt{AV}$-$\mathcal{L}_{P}^{\text{CD}}$} &2.03 &6.08 &4.46 &6.09  \\ 
\hline
\hline
\small{$\mathtt{RE}$-$\mathcal{L}_{P}^{\text{CD}}$, $\mathtt{var.}$ $\mathtt{IoR}$}& 2.13 & 6.24 & 4.56 & 6.11 \\ 
\hline
\hline
PSR & 5.13 & 6.94 & 4.75 & 14.7 \\
\hline
\end{tabular}
\vspace{0.1cm}
\caption{Quantitative comparisons of point cloud reconstruction from 10 views. $\mathtt{RE}$, $\mathtt{NE}$ and $\mathtt{AV}$ represent feature mapping methods: rendering error based view selection, nearest view selection and average fusion, respectively. $\mathcal{L}_{P}^{\text{nearest}}$, $\mathcal{L}_{P}^{\text{view}}$ and $\mathcal{L}^{\text{CD}}_{P}$ are the loss functions defined in Sec.~\ref{sec:method}. $\mathtt{RE}$-$\mathcal{L}_{P}^{\text{CD}}$, $\mathtt{var.}$ $\mathtt{IoR}$ represents sensitivity analysis for IoR, as described in text. PSR represents optimization \cite{kazhdan2006poisson} to refine the point cloud based on predicted normals. }
\label{tab:pointQuan}
\vspace{-0.2cm}
\end{table}

\begin{figure*}
\centering
\includegraphics[width=0.98\textwidth]{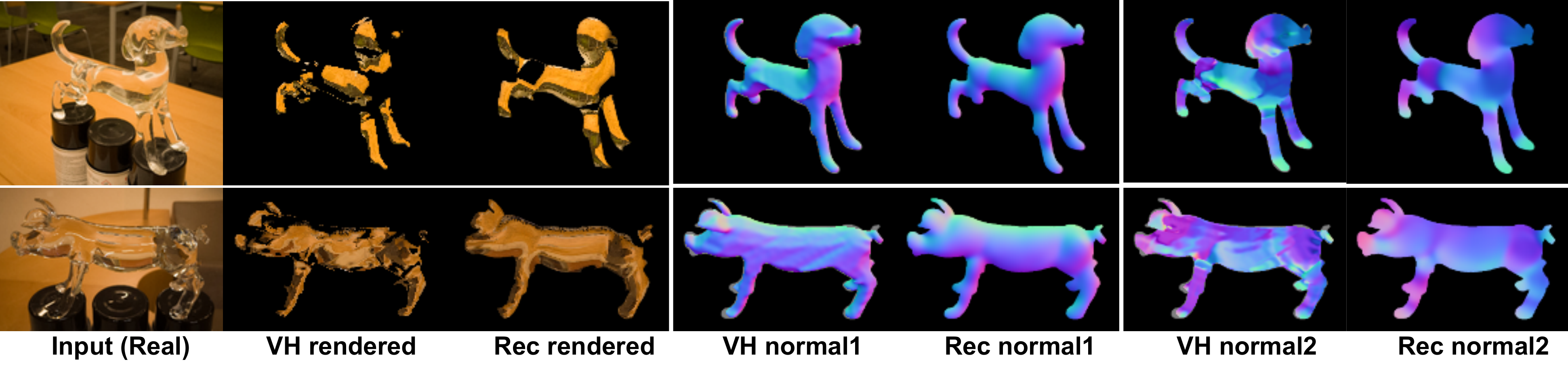}
\vspace{-0.2cm}
\caption{Normal reconstruction of real transparent objects and the rendered images. The initial visual hull normals are built from 10 views. The region of total reflection has been masked out in the rendered images. }
\label{fig:real_normal}
\end{figure*}

\begin{figure*}
%\vspace{-0.2cm}
\centering
\includegraphics[width=0.98\textwidth]{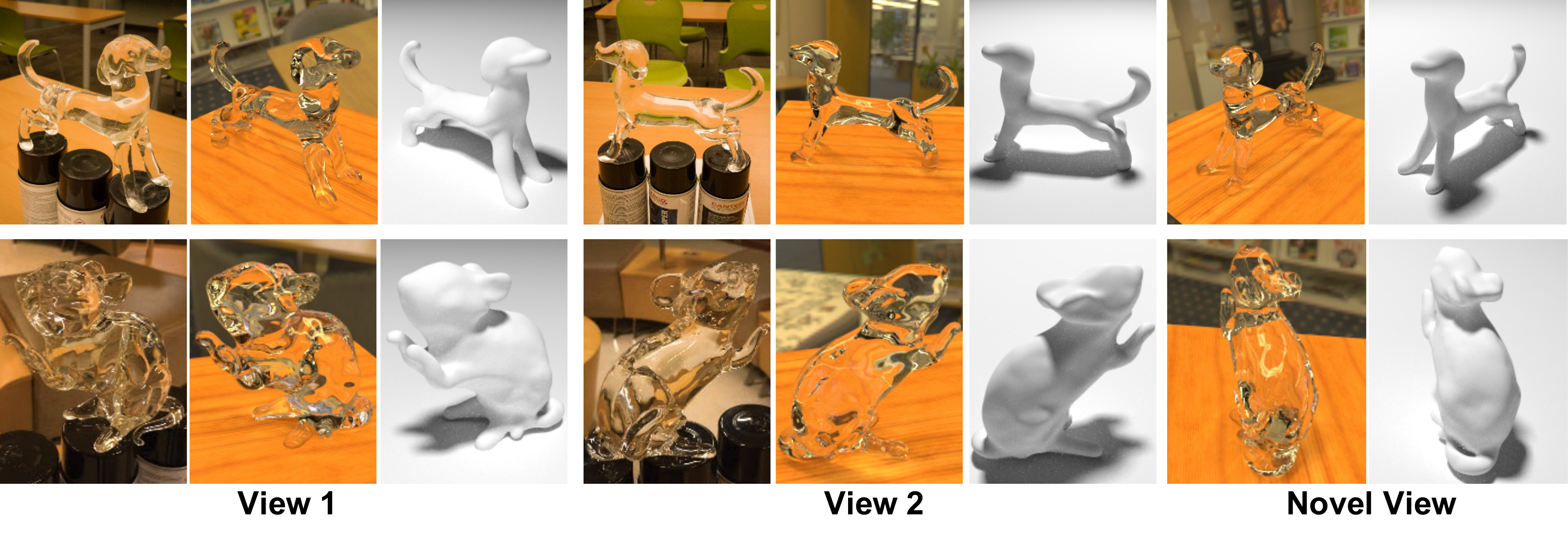}
\vspace{-0.2cm}
\caption{3D shape reconstruction on real data. Columns 1-6 show reconstruction results from 2 known view directions. For each view, we show the input image and the reconstructed shape rendered from the same view under different lighting and materials. Columns 7-8 render the reconstructed shape from a novel view direction that has not been used to build the visual hull. The first shape is reconstructed using only 5 views (top row) while the second uses 10 views (bottom row). Also see comparisons to ground truth scans in supplementary material. }
\label{fig:real_multi_views}
\vspace{-0.2cm}
\end{figure*}

\begin{figure}
\centering
\vspace{-0.2cm}
\includegraphics[width=3.25in]{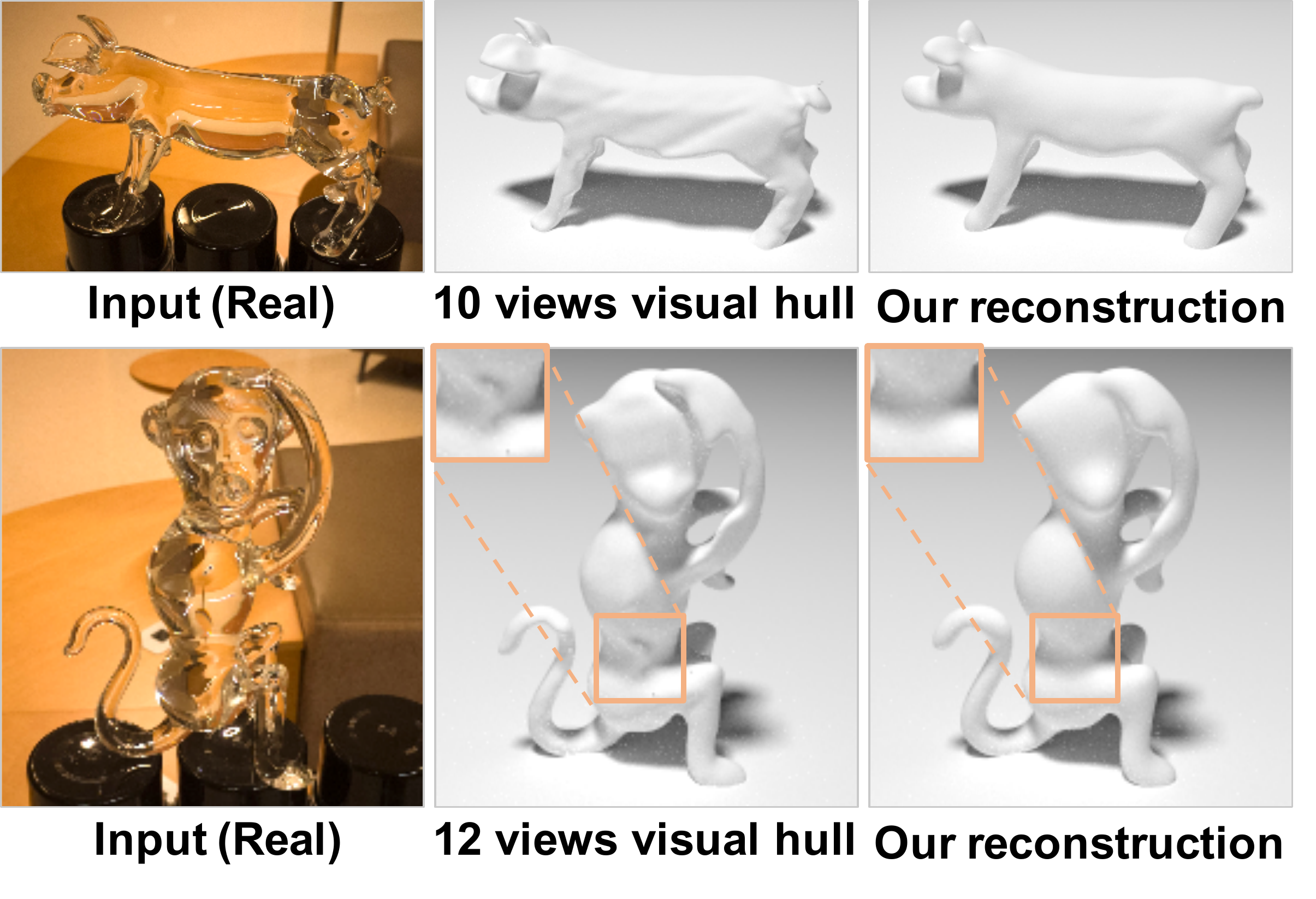}
\vspace{-0.4cm}
\caption{Comparison between visual hull initialization and our shape reconstruction on real objects. Our method recovers more details, especially for concave regions. }
\label{fig:vhComparion}
\vspace{-0.5cm}
\end{figure}

\vspace{-0.4cm}
\paragraph{Point cloud reconstruction} Quantitative comparisons of the 10-view point cloud reconstruction network are summarized in Table \ref{tab:pointQuan}. After obtaining the point and normal predictions $\{p\}$ and $\{N\}$, we reconstruct 3D meshes as described above. We compute the Chamfer distance (CD), Chamfer normal median angle (CDN-med), Chamfer normal mean angle (CDN-mean) and Metro distance by uniformly sampling 20000 points on the ground-truth and reconstructed meshes. We first compare the effectiveness of different loss functions. We observe that while all the three loss functions can greatly improve the reconstruction accuracy compared with the initial 10-view visual hull, the Chamfer distance loss ($\mathtt{RE}$-$\mathcal{L}_{P}^{\text{CD}}$) performs significantly better than view-dependent loss ($\mathtt{RE}$-$\mathcal{L}_{P}^{\text{view}}$) and nearest $L_2$ loss ($\mathtt{RE}$-$\mathcal{L}_{P}^{\text{nearest}}$). %which suggests that Chamfer distance loss is a better metric compared with the other two losses to measure differences between 2 point clouds. 
Next, we test different feature mapping strategies, where the rendering error based view selection method ($\mathtt{RE}$-$\mathcal{L}_{P}^{\text{CD}}$) performs consistently better than the other two methods. This is because our rendering error can be used as a meaningful metric to predict normal reconstruction accuracy, which leads to better point cloud reconstruction.  Ablation studies for the modified PointNet++ are included in supplementary material. 

The last row of Table \ref{tab:pointQuan} shows that an optimization-based method like PSR \cite{kazhdan2006poisson} to refine shape from predicted normals does not lead to much improvement, possibly since visual hull shapes are still significantly far from ground truth. In contrast, our network allows large improvements.

\begin{table}[t]
\footnotesize
\renewcommand{\arraystretch}{1.2}
\setlength{\tabcolsep}{2pt}
\centering
\begin{tabular}{l|c|c|c|c}
\hline
& CD($10^{-4}$) & CDN-mean($^\circ$) & CDN-med($^\circ$) & Metro($10^{-3}$) \\
\hline
$\mathtt{vh5}$ &31.7 &13.1 &10.3 &66.6 \\
%\hline
$\mathtt{Rec5}$ &\textbf{6.30} &\textbf{11.0} & \textbf{8.7} & \textbf{15.2} \\
\hline
\hline 
$\mathtt{vh20}$ & 2.23 &4.59 & \textbf{2.71} & 6.83 \\
%\hline
$\mathtt{Rec20}$ & \textbf{1.20} & \textbf{4.04} &2.73 & \textbf{4.18} \\
\hline
\end{tabular}
\vspace{0.1cm}
\caption{Quantitative comparisons of point cloud reconstruction from 5 views and 20 views. In both cases, our pipeline significantly improves the transparent shape reconstruction accuracy compared with classical visual hull method. }
\label{tab:pointViewsQuan}
\vspace{-0.3cm}
\end{table}

\vspace{-0.4cm}
\paragraph{Different number of views} We also test the entire reconstruction pipeline for 5 and 20 views, with results summarized in Table \ref{tab:pointViewsQuan}. We use the setting that leads to the best performance for 10 views, that is, $\mathtt{wr}+\mathtt{cv}+\mathtt{op}$ for normal reconstruction and $\mathtt{RE}$-$\mathcal{L}_{P}^{\text{CD}}$ for point cloud reconstruction, achieving significantly lower errors than the visual hull method. Figure \ref{fig:synDiffView} shows an example from the synthetic test set for reconstructions with different number of views. Further results and comparisons are in supplementary material. 

\vspace{-0.4cm}
\paragraph{Sensitivity analysis for IoR} 
We also evaluate the model on another test set with the same geometries, but unknown IoRs sampled uniformly from the range $\left[ 1.3, 1.7 \right]$. As shown in Tables \ref{tab:normalQuan} and \ref{tab:pointQuan}, errors increase slightly but stay reasonable, showing that our model can tolerate inaccurate IoRs to some extent. Detailed analysis is in the supplementary material.

\subsection{Results on Real Transparent Objects}
\vspace{-0.2cm}
We acquire RGB images using a mobile phone.
%with a similar field of view as our rendering. 
To capture the environment map, we take several images of a mirror sphere at the same location as the transparent shape. We use COLMAP~\cite{schoenberger2016sfm} to obtain the camera poses and manually create the segmentation masks.
%used along with the environment maps and images to generate the 3D mesh. 
%Note that the IORs of our real transparent shapes are the same as our synthetic dataset. 
\vspace{-0.4cm}
\paragraph{Normal reconstruction}
We first demonstrate the normal reconstruction results on real transparent objects in Figure~\ref{fig:real_normal}. Our model significantly improves visual hull normal quality. The images rendered from our predicted normals are much more similar to the input RGB images compared to those rendered from visual hull normals. %More qualitative results are shown in the supplementary materials.

\vspace{-0.4cm}
\paragraph{3D shape reconstruction}
In Figure~\ref{fig:real_multi_views}, we demonstrate our 3D shape reconstruction results on real world transparent objects under natural environment map. The dog shape in the first row only takes 5 views and the mouse shape in the second row takes 10 views. We first demonstrate the reconstructed shape from the same view as the input images by rendering them under different lighting and materials. Even with very limited inputs, our reconstructed shapes are still of high quality. To test the generalizability of our predicted shapes, we render them from novel views that have not been used as inputs and the results are still reasonable. Figure \ref{fig:vhComparion} compares our reconstruction results with the visual hull initialization. We observe that our method performs much better, especially for concave regions. Comparisons with scanned ground truth are in supplementary material. 

\vspace{-0.4cm}
\paragraph{Runtime} Our method requires around 46s to reconstruct a transparent shape from 10 views on a 2080 Ti, compared to 5-6 hours for previous optimization-based methods \cite{wu2018full}.

\vspace{-0.2cm}
\section{Discussion}
\label{sec:discussion}
\vspace{-0.2cm}

We present the first physically-based deep network to reconstruct transparent shapes from a small number of views captured under arbitrary environment maps. 
Our network models the properties of refractions and reflections through a physically-based rendering layer and cost volume, to estimate surface normals at both the front and back of the object, which are used to guide a point cloud reconstruction. Extensive experiments on real and synthetic data demonstrate that our method can recover high quality 3D shapes.  

\vspace{-0.4cm}
\paragraph{Limitations and future work} 
Our limitations suggest interesting future avenues of research. A learnable multiview fusion might replace the visual hull initialization. We believe more complex light paths of length greater than $3$ may be handled by differentiable path tracing along the lines of differentiable rendering \cite{li2018differentiable,zhang2019DTRT}. While we assume a known refractive index, it may be jointly regressed. Finally, since we reconstruct $N^2$, future works may also estimate the back surface to achieve single-view 3D reconstruction. 

\vspace{-0.4cm}
\paragraph{Acknowledgments}
This work is supported by NSF CAREER Award 1751365 and a Google Research Award. We thank Adobe and Cognex for generous support. We thank Hui Huang and Bojian Wu for their help on obtaining the real transparent shapes.

\appendix
\section{Real Data Evaluation with Ground Truth}
\label{sec:moreRrealResults}

We first present quantitative comparisons and then more qualitative ones on real data. We use four objects ({\em mouse}, {\em dog}, {\em pig} and {\em monkey}). All objects are reconstructed from 10 views under natural environment maps, except {\em monkey}, which needs 12 views since the shape is much more complex. To obtain the ground truth geometry, we paint each object with diffuse white paint and scan using a high-quality 3D scanner. All code and data will be publicly released.

\vspace{-0.3cm}
\paragraph{Quantitative results}
We manually align ground-truth shapes with the predicted shapes using ICP method \cite{besl1992method} and then uniformly sample 20000 points on the both shapes to compute the four error metrics (CD, CDN-mean, CDN-med, Metro). The quantitative numbers are summarized in Table~\ref{tab:pointQuanReal}. For all the 4 objects, our method consistently outperforms the visual hull baseline, which again demonstrates the effectiveness of our transparent shape reconstruction framework.

% \begin{table}[h]
% \footnotesize
% \renewcommand{\arraystretch}{1.2}
% \setlength{\tabcolsep}{1pt}
% \centering
% \begin{tabular}{l|c|c|c|c|c|c|c|c|c}
% \hline
% & Views & \multicolumn{2}{|c|}{CD($10^{-4}$)} & \multicolumn{2}{|c|}{CDN-mean($^\circ$)} & \multicolumn{2}{|c|}{CDN-med($^\circ$)} & \multicolumn{2}{|c}{Metro($10^{-3}$)} \\
% \hline
% & & $~~~~\mathtt{vh}~~~~$ & $~~\mathtt{Rec}~~$ & $~~~\mathtt{vh}~~~$ & $\mathtt{Rec}$ & $~~~\mathtt{vh}~~~$ & $\mathtt{Rec}$ & $~~~\mathtt{vh}~~~$ & $\mathtt{Rec}$ \\
% \hline
% monkey &12 & 3.99 & \textbf{3.94} & 21.2 & \textbf{16.4} & 14.8 & \textbf{11.9} & 20.7 & \textbf{13.9}  \\
% mouse &10 &8.29 & \textbf{5.67} & 21.5 & \textbf{18.5} & 15.6 & \textbf{13.9} & 16.9 & \textbf{14.1}  \\
% pig &10 &5.64 & \textbf{4.95} & 20.4 & \textbf{19.7} & \textbf{14.7} & 15.5 & 13.0 & \textbf{7.53}  \\
% dog &10 &2.38 & \textbf{1.98}  & 15.8 & \textbf{13.7} & 12.1 & \textbf{11.1} & 4.86 & \textbf{4.80}  \\
% \hline
% mean &10.5 &5.08 & \textbf{4.14}  & 19.7 & \textbf{17.1} & 14.3 & \textbf{13.1} & 13.9 & \textbf{10.1}  \\
% \hline
% \end{tabular}
% \vspace{0.05cm}
% \caption{Quantitative comparisons of transparent shape reconstruction on real data. We observe that our reconstruction achieves lower average errors than the  visual hull method on all the metrics.}
% \label{tab:pointQuanReal}
% \vspace{-0.3cm}
% \end{table}
\begin{table}[h]
\footnotesize
\renewcommand{\arraystretch}{1.2}
\setlength{\tabcolsep}{1pt}
\centering
\begin{tabular}{l|c|c|c|c|c|c|c|c|c}
\hline
& Views & \multicolumn{2}{|c|}{CD($10^{-4}$)} & \multicolumn{2}{|c|}{CDN-mean($^\circ$)} & \multicolumn{2}{|c|}{CDN-med($^\circ$)} & \multicolumn{2}{|c}{Metro($10^{-3}$)} \\
\hline
& & $~~~~\mathtt{vh}~~~~$ & $~~\mathtt{Rec}~~$ & $~~~\mathtt{vh}~~~$ & $\mathtt{Rec}$ & $~~~\mathtt{vh}~~~$ & $\mathtt{Rec}$ & $~~~\mathtt{vh}~~~$ & $\mathtt{Rec}$ \\
\hline
monkey &12 & 3.99 & \textbf{3.94} & 21.2 & \textbf{16.4} & 14.8 & \textbf{11.9} & 20.7 & \textbf{13.9}  \\
mouse &10 &8.04 & \textbf{5.35} & 19.0 & \textbf{16.3} & 11.4 & \textbf{12.0} & 16.6 & \textbf{13.0}  \\
pig &10 &5.58 & \textbf{4.87} & 19.0 & \textbf{18.3} & \textbf{14.0} & 14.6 & 13.0 & \textbf{7.4}  \\
dog &10 &2.25 & \textbf{1.86}  & 14.5 & \textbf{12.4} & 11.4 & \textbf{10.3} & 4.1 & \textbf{4.0}  \\
\hline
mean &10.5 &4.97 & \textbf{4.00}  & 18.4 & \textbf{15.9} & 12.9 & \textbf{12.2} & 13.6 & \textbf{9.6}  \\
\hline
\end{tabular}
\vspace{0.05cm}
\caption{Quantitative comparisons of transparent shape reconstruction on real data. We observe that our reconstruction achieves lower average errors than the  visual hull method on all the metrics.}
\label{tab:pointQuanReal}
\vspace{-0.3cm}
\end{table}

\begin{figure*}
\centering
\includegraphics[width=0.97\textwidth]{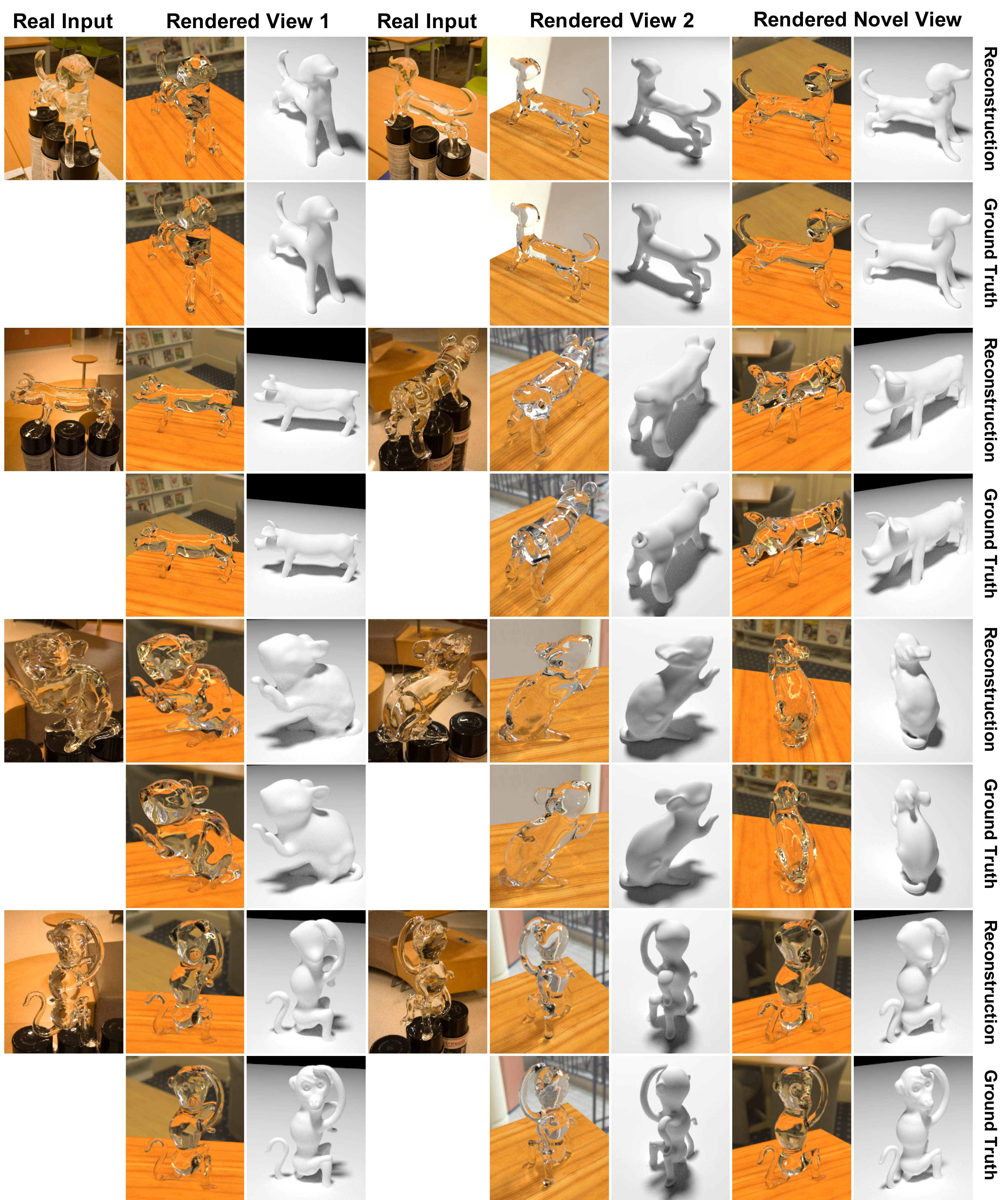}
\vspace{-0.1cm}
\caption{Results on 3D reconstruction for four real transparent objects. All shapes are reconstructed from 10 views, except the {\em monkey} in the last row that uses 12 views. We first present reconstruction results from two input views (columns 1-6). From left to right, the odd rows show the input image and the reconstructed shapes under different lighting and materials. The corresponding outputs using the ground-truth shapes rendered from the same view are shown in the even rows. We also render the reconstructed shapes and ground-truth shapes from a novel view direction that has not been used to build the visual hull (columns 7-8). In each instance, we observe that the reconstructions are close to the ground truth despite the challenging shapes, complex light paths and small number of views used for 3D reconstruction.}
\label{fig:realData}
\end{figure*}

\vspace{-0.3cm}
\paragraph{Qualitative results and videos}
Figure \ref{fig:realData} shows both the ground-truth transparent shapes and our reconstructed shapes rendered under different lighting and materials. Even though the shapes are complex and we use very limited inputs, our reconstructions still closely match the ground truth appearance. This demonstrates the efficacy of our physically-motivated network that models complex light paths induced by refractions and reflections. To better visualize the quality of our 3D reconstruction outputs, we create a video by rotating both the ground-truth shapes and the reconstructed shapes under different natural environment maps. The video is included in the supplementary material. A higher resolution video is available at this \href{https://drive.google.com/file/d/1bkcRYg55WkaQuVXI3PDAmQIwmk4Y0Tf1/view?usp=sharing}{link}.

\begin{figure*}
\centering
\includegraphics[width=\textwidth]{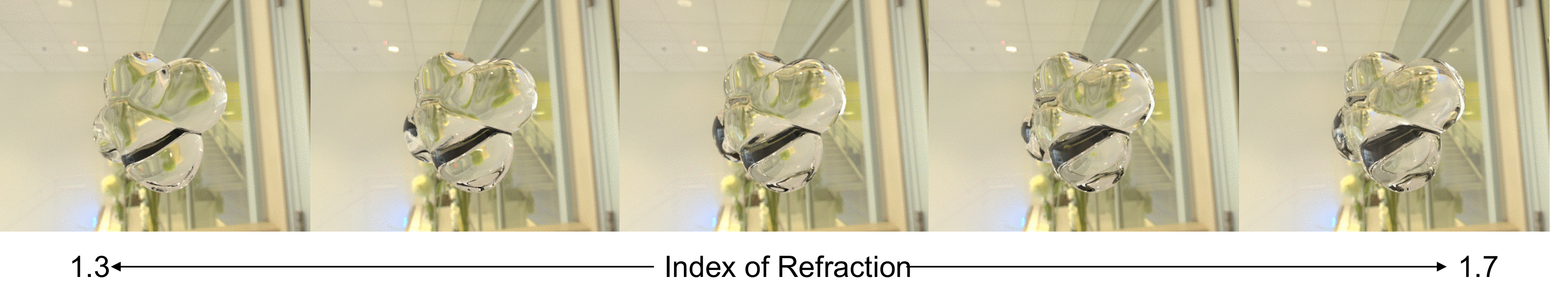}
\vspace{-0.5cm}
\caption{Appearance changes for same shape geometry under various index of refraction (IoRs). IoRs range from 1.3 to 1.7.}
\label{fig:iors}
\end{figure*}

\section{Sensitivity Analysis for Index of Refraction}
\label{sec:sen_ior}
As mentioned in Sec. 4.1 of the main paper, we perform a sensitivity analysis on the influence of a different test-time IoR on the shape reconstruction accuracy. We re-render our synthetic transparent testing set with the same shapes and environment maps. However, instead of rendering with a fixed IoR value of 1.4723, we randomly sample 5 different IoRs ranging from $1.3$ to $1.7$ for each shape. Figure \ref{fig:iors} shows an example of the same shape rendered under different IoRs. We then test our network trained with a fixed IoR value of 1.4723 on the new test set with variable IoR. During testing, the IoR used by the rendering layer is kept fixed at 1.4723. The quantitative comparisons have been summarized in Tables 1 and 2 of the main paper. 

\begin{figure}
\centering
\includegraphics[width=0.5\textwidth]{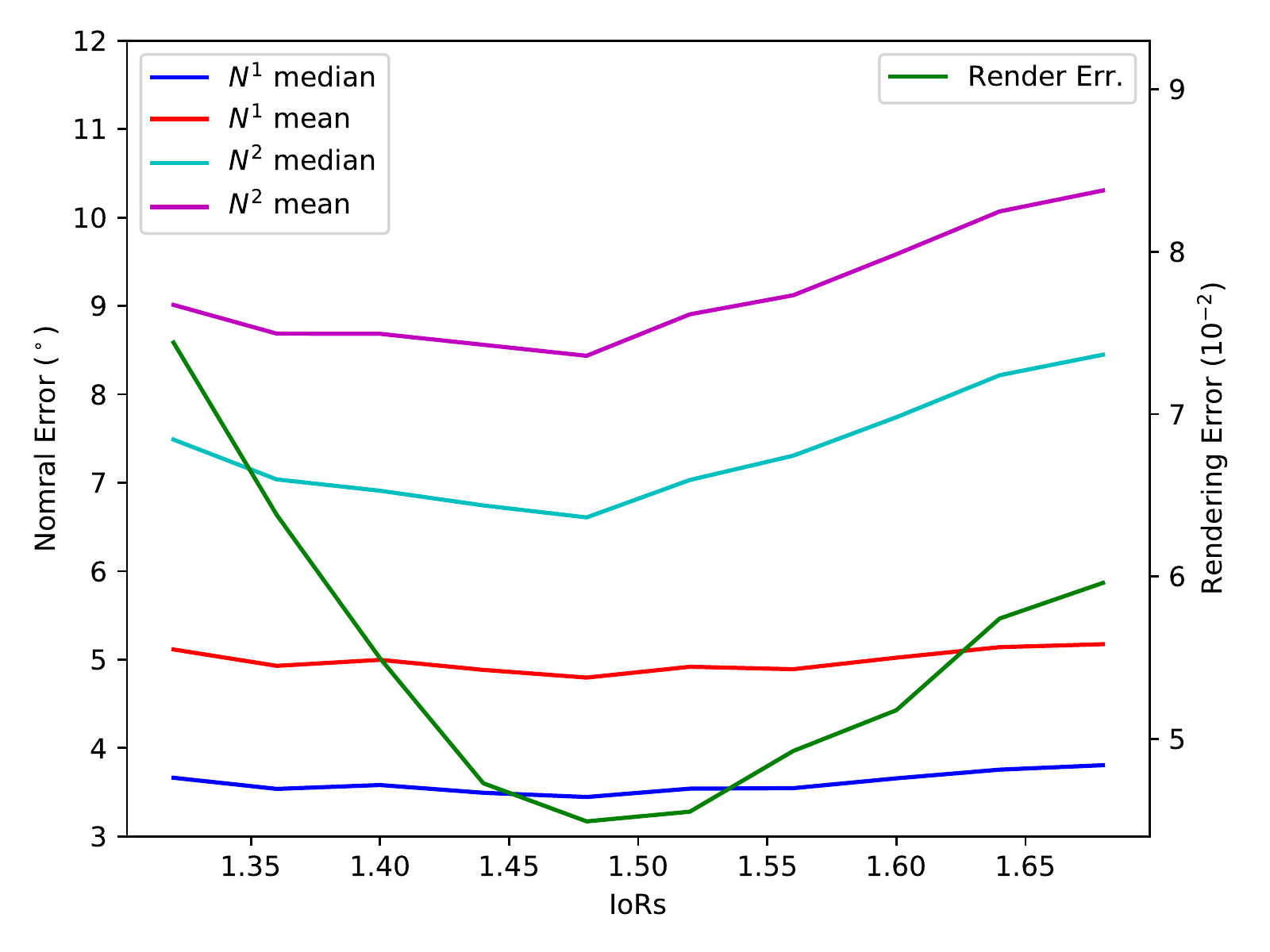}
\vspace{-0.5cm}
\caption{The mean normal estimation errors across varying IoRs in the test set, using the fixed training set IoR value for prediction.}
\vspace{-0.3cm}
\label{fig:ior_normal}
\end{figure}

\begin{figure}
\centering
\includegraphics[width=0.5\textwidth]{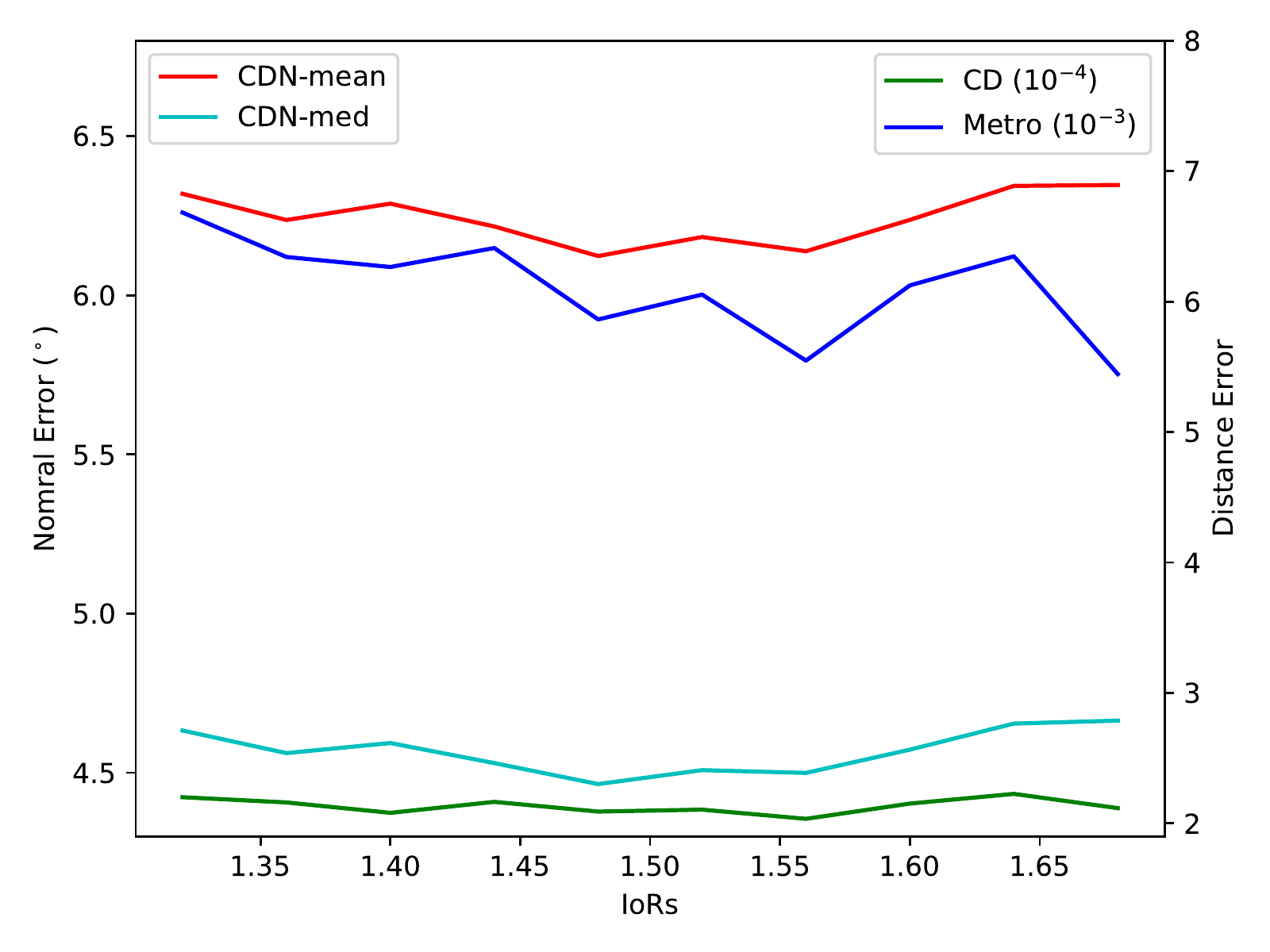}
\vspace{-0.5cm}
\caption{The mean shape reconstruction errors across varying IoRs in the test set, using the fixed training set IoR value for prediction.}
\vspace{-0.3cm}
\label{fig:ior_point}
\end{figure}

Figure~\ref{fig:ior_normal} and~\ref{fig:ior_point}  show trends in the normal and shape reconstruction errors across varying IoRs in the test set. As expected, the errors are relatively smaller for IoRs close to the training set value of $1.4723$. In particular, this trend is more explicitly visible in normal estimation, since the model leverages the features from the rendering layer and cost volume which require known IoRs. However, the overall variation in error is small across this range of IoRs. 

The above plots further support the analysis in Tables 1 and 2 of the main paper. Even though the predicted normals and the final reconstructed mesh are expectedly more accurate in the known IoR case, the quantitative errors increase gracefully and not too much across a range even with unknown IoR. This suggests that our network is relatively robust to the IoR value. As stated in the main paper, our future work will consider simultaneously reconstructing the transparent shape and predicting its IoR. 

%As mentioned in Sec. 4.1, we had sensitivity analysis on IoRs. We rendered another synthetic transparent dataset which contains same shape geometries as ones in the original dataset. The testing set for the new dataset contains exactly the same geometries with corresponding environment maps as the ones in the original testing set, but with five different IoRs sampled from $(1.3, 1.7)$ instead of $1.4723$ for each shape. Index of Refraction (IoR) plays an important role in our differential rendering layer. Given any shape geometry, the appearance of transparent shape will be very different under different IoRs, as observed in Figure~\ref{fig:iors}. As our differentiable rendering layer assumes a fixed IoR, the rendered images for input whose IoR is not $1.4723$ will not provide the informative visual clues for our model to build cost volume and compute rendering error. Therefore, the performance on inputs with various IoRs is slightly worse than inputs with fixed IoR. 

\section{Further Ablation Studies}
\label{sec:ablationStudies}

\begin{figure*}
\centering
\includegraphics[width=6.4in]{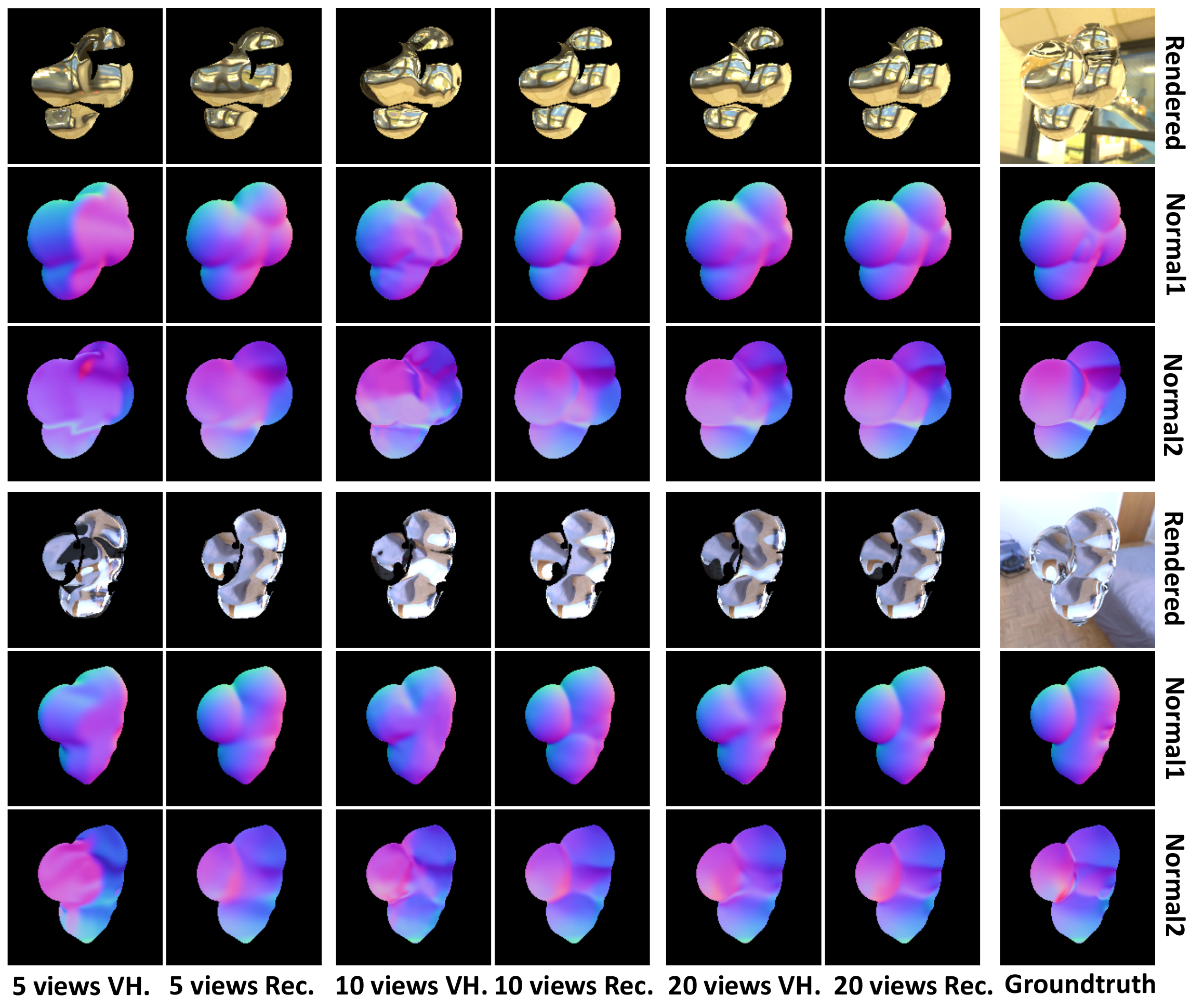}
\caption{Normal predictions on our synthetic dataset with different number of input views. Regions with total reflection have been masked out in the rendered images. Our predicted normals are much closer to the ground truth compared to the visual hull normals.}
\vspace{-0.3cm}
\label{fig:diffViewNormal}
\end{figure*}

\begin{figure*}
\centering
\includegraphics[width=6.4in]{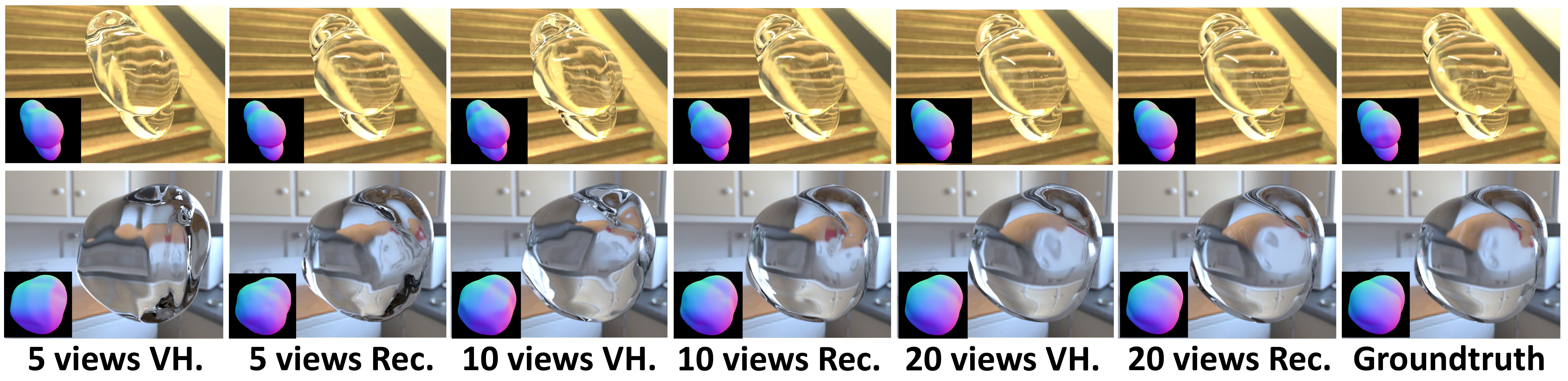}
\caption{Transparent shape reconstruction in our synthetic dataset using 5, 10 and 20 views. Images rendered with our reconstructed shapes are much closer to the those rendered with ground truth shape, as compared to images rendered with the visual hull shapes. The inset normals are rendered from the reconstructed shapes and demonstrate the same conclusion.}
\vspace{-0.3cm}
\label{fig:diffViewShape}
\end{figure*}

\begin{table}[t]
\footnotesize
\renewcommand{\arraystretch}{1.2}
\setlength{\tabcolsep}{4pt}
\centering
\begin{tabular}{l|c|c|c|c|c}
\hline
5 views normal & \multirow{2}{*}{$\mathtt{vh5}$} & \multirow{2}{*}{$\mathtt{basic}$}  & \multirow{2}{*}{$\mathtt{wr}$} & \multirow{2}{*}{$\mathtt{wr}$+$\mathtt{cv}$} & $\mathtt{wr}$+$\mathtt{cv}$  \\
reconstruction & & & & & +$\mathtt{op}$ \\
\hline
$N^1$ median ($^\circ$) &12.7 & 6.1 & 6.0 & 6.0 & \textbf{5.9} \\
\hline
$N^{1}$ mean ($^\circ$) &15.3 & 7.8 & 7.9 & 7.8 & \textbf{7.7} \\
\hline 
$N^2$ median ($^\circ$) &18.3 & 10.7 & 10.7 & 10.5 & \textbf{10.0} \\
\hline 
$N^{2}$ mean ($^\circ$) &20.9 & 12.5 & 12.5 & 12.3 & \textbf{11.9} \\
\hline
Render Err.($10^{-2}$) &9.7 & 5.9 & 5.8 & 5.9 & \textbf{4.1} \\
\hline
\hline
20 views normal & \multirow{2}{*}{$\mathtt{vh20}$} & \multirow{2}{*}{$\mathtt{basic}$}  & \multirow{2}{*}{$\mathtt{wr}$} & \multirow{2}{*}{$\mathtt{wr}$+$\mathtt{cv}$} & $\mathtt{wr}$+$\mathtt{cv}$  \\
reconstruction & & & & & +$\mathtt{op}$ \\
\hline
$N^1$ median ($^\circ$) &2.5 & 2.2 & 2.2 & 2.2 & \textbf{2.2} \\
\hline
$N^{1}$ mean ($^\circ$) &4.6 & 3.4 & 3.4 & 3.3 & \textbf{3.3} \\
\hline 
$N^2$ median ($^\circ$) &5.2 & 4.7 & 4.6 & 4.6 & \textbf{4.3} \\
\hline 
$N^{2}$ mean ($^\circ$) &7.6 & 6.5 & 6.4 & 6.3 & \textbf{6.1} \\
\hline
Render Err.($10^{-2}$) &4.0 & 3.7 & 3.8 & 3.8 & \textbf{2.7} \\
\hline
\end{tabular}
\vspace{0.1cm}
\caption{Quantitative comparisons of normal estimation from 5 and 20 views. Following the notation in the main paper, $\mathtt{vh5}$ and $\mathtt{vh20}$ represent the initial normals reconstructed from visual hulls corresponding to 5 and 20 views, respectively. Here,  $\mathtt{wr}$ and $\mathtt{basic}$  are our basic encoder-decoder network with and without rendering error map ($I^{er}$) and total reflection mask ($M^{tr}$) as inputs. Further, $\mathtt{wr}$+$\mathtt{cv}$ represents our network with cost volume and  $\mathtt{wr}$+$\mathtt{cv}$ + $\mathtt{opt}$ represents the predictions after optimizing the latent vector to minimize the rendering error. Similar to the 10-view case, $\mathtt{wr}$+$\mathtt{cv}$ + $\mathtt{opt}$ performs better than all other baselines for transparent shape reconstruction using both 5 and 20 views.}
\vspace{-0.3cm}
\label{tabsup:normalQuan}
\end{table}

\vspace{-0.1cm}
\paragraph{Different number of views} Table \ref{tabsup:normalQuan} summarizes the normal predictions from 5 and 20 views. Similar to the 10-view case, our entire method $\mathtt{wr}$+$\mathtt{cv}$+$\mathtt{op}$ outperforms all other baselines on all the five metrics. In particular, we find the cost volume ($\mathtt{cv}$) and the optimization of the latent vector ($\mathtt{op}$) bring the largest improvements on normal reconstruction accuracy. This justifies our intuition that utilizing the correspondence between the input image and the captured environment map by modeling the image formation process within the network can lead to better normal reconstruction results. Figure \ref{fig:diffViewNormal} shows two normal reconstruction results on our synthetic dataset. For both examples, our physically-based network performs significantly better than the classical visual hull method for 5, 10 and 20 views. 

\begin{table}[t]
\footnotesize
\renewcommand{\arraystretch}{1.2}
\setlength{\tabcolsep}{1pt}
\centering
\begin{tabular}{l|c|c|c|c}
\hline
& CD($10^{-4}$) & CDN-mean($^\circ$) & CDN-med($^\circ$) & Metro($10^{-3}$) \\
\hline
\small{$\mathtt{RE}$-$\mathcal{L}_{P}^{\text{CD}}$}& \textbf{2.00} & \textbf{6.02} & \textbf{4.38} & \textbf{5.98} \\ 
\hline
\hline
--~ maxPooling &2.09 & 6.26 &4.59 & 6.09\\ 
--~ normal Diff.        &2.09 &6.31 &4.62 &6.48\\
--~ normal Skip.        &2.07 &6.14 &4.51 & 6.20 \\
standard & 2.12 & 6.34 & 4.72 & 6.49 \\
\hline
\end{tabular}
\vspace{0.1cm}
\caption{Comparisons of point cloud reconstruction with different PointNet++ architectures on our synthetic dataset. Following the notation in the main paper, $\mathtt{RE}$ represents rendering error based view selection. $\mathcal{L}^{\text{CD}}_{P}$ represents the Chamfer distance loss. }
\label{tabsup:pointQuan}
\vspace{-0.3cm}
\end{table}

\begin{figure}
\centering
\includegraphics[width=0.49\textwidth]{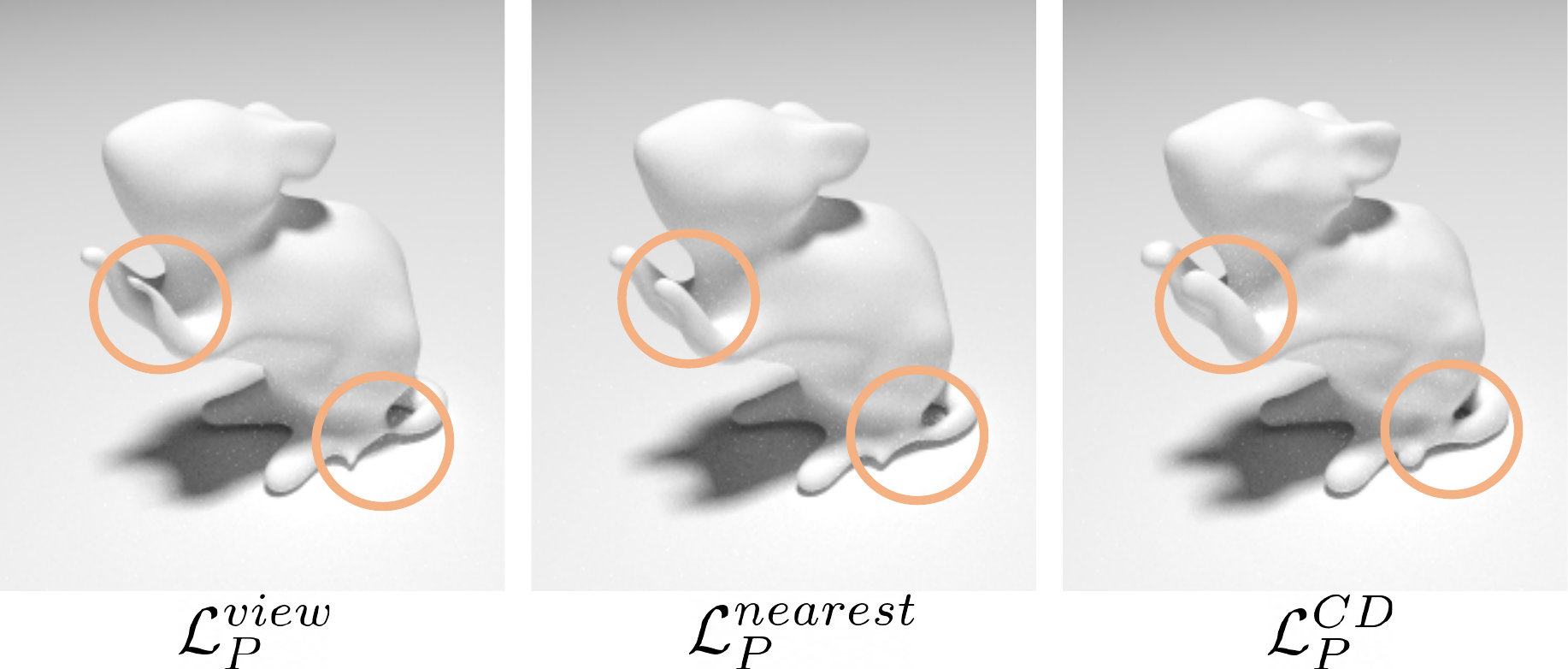}
\caption{Comparisons of point cloud reconstruction with different loss functions on a real example. Our modified PointNet++ trained with Chamfer distance loss achieves better quality compared with the other two losses. }
\label{fig:pointQualiReal}
\end{figure}

\vspace{-0.3cm}
\paragraph{Modification of standard PointNet++ \cite{qi2017pointnet++}} 
We examine our modifications of the standard PointNet++ architecture for point cloud reconstruction to better incorporate normal information. The quantiative numbers are summarized in Table \ref{tabsup:pointQuan}. We first remove single modifications from standard PointNet++ to our novel version (--~ maxPooling, --~normal Diff.~and --~normal Skip.) and then remove all the modifications to use standard PointNet++ to reconstruct the point cloud of our transparent shapes (standard). Experiments show that each of our modifications brings consistent improvements in reconstruction accuracy and removing all of them leads to a much poorer performance. This shows our modifications ease the difficulty for the network to reason about point cloud distribution based on normal predictions. Figure \ref{fig:pointQualiReal} demonstrates a real example reconstructed by our modified PointNet++ trained using different loss functions. It is clearly observed that our modified PointNet++ trained with Chamfer distance loss leads to a more complete and less noisy 3D reconstruction, especially for thin structures and concave regions. 

\begin{table}[t]
\footnotesize
\renewcommand{\arraystretch}{1.2}
\setlength{\tabcolsep}{4pt}
\centering
\begin{tabular}{l|c|c|c}
\hline
10 views normal & \multirow{2}{*}{$\mathtt{vh10}$} & $\mathtt{wr}$+$\mathtt{cv}$ &  $\mathtt{wr}$+$\mathtt{cv}$\\
reconstruction & &  +$\mathtt{op}$ &  +$\mathtt{opPixel}$ \\
\hline
$N^1$ median ($^\circ$) & 5.5  & \textbf{3.4}  & 3.8 \\
\hline
$N^{1}$ mean ($^\circ$) & 7.5  & \textbf{4.8}  & 4.9 \\
\hline 
$N^2$ median ($^\circ$) & 9.2  & \textbf{6.6}  & 7.4 \\
\hline 
$N^{2}$ mean ($^\circ$) & 11.6  & \textbf{8.4}  & 8.5 \\
\hline
Render Err.($10^{-2}$) & 6.0  & 2.9 & \textbf{2.6}  \\
\hline
\end{tabular}
\vspace{0.05cm}
\caption{Quantitative comparisons of different optimization strategies for normal estimation from 10 views. $\mathtt{op}$ represents optimization the latent vector, which is the results reported in the main paper. $\mathtt{opPixel}$ represents optimization direction in the pixel space. }
\vspace{-0.3cm}
\label{tabsup:normalQuanOpt}
\end{table}

\vspace{-0.3cm}
\paragraph{Optimization of latent vector} 
We adopt an alternating minimization strategy to optimize the latent vector. We first keep $N^{1}$ unchanged and only change $N^2$ by adding a large identity loss on $N^1$. After 500 iterations, we remove the constraint and optimize both $N^1$ and $N^2$ simultaneously. This is because the our $N^1$ prediction is usually more accurate and optimizing $N^2$ first can lead to better results. In Table \ref{tabsup:normalQuanOpt}, we compare the normal reconstruction results of optimizing the latent vector and directly optimizing the per-pixel normals. The quantitative comparison shows that while optimizing per-pixel normal can also decrease the rendering error, only by optimizing the latent vector can we observe  improvements in normal reconstruction accuracy. The inherent ill-posed nature of normal prediction of transparent shapes makes it necessary to have a strong regularization to obtain meaningful outputs. In this case, the regularization is provided by the trained decoder which constrains the predicted normals to be on the natural shape manifold. 
\section{Building the Cost Volume}
\label{sec:cost_volume}

\begin{table}[t]
\footnotesize
\renewcommand{\arraystretch}{1.1}
\centering
\begin{tabular}{|l|c|c|}
\hline
& $\{\theta_{k}\}_{k=1}^{4}$ & $\{\phi_{k}\}_{k=1}^{4}$ \\
\hline
5 views & $0^{\circ}$,$25^{\circ}$,$25^{\circ}$,$25^{\circ}$ & $0^{\circ}$,$0^{\circ}$,$120^{\circ}$,$240^{\circ}$  \\
 \hline
10 views & $0^{\circ}$,$15^{\circ}$,$15^{\circ}$,$15^{\circ}$  & $0^{\circ}$,$0^{\circ}$,$120^{\circ}$,$240^{\circ}$   \\
\hline
20 views & $0^{\circ}$,$10^{\circ}$,$10^{\circ}$,$10^{\circ}$ &$0^{\circ}$,$0^{\circ}$,$120^{\circ}$,$240^{\circ}$ \\
\hline
\end{tabular}
\vspace{0.1cm}
\caption{The sampled angles for building cost volume. We set the sampled angles according to the normal error of visual hull reconstructed by different number of views. }
\label{tab:sampledAngles} 
\vspace{-0.3cm}
\end{table}

\begin{figure}
\centering
\includegraphics[width=3.3in]{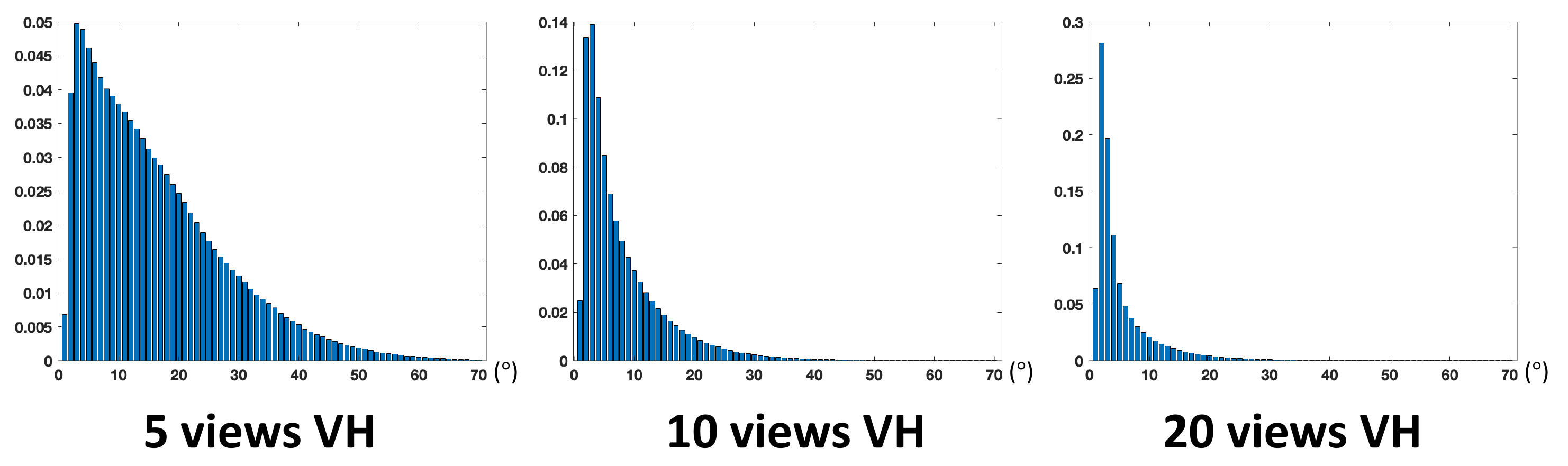}
\caption{The error distribution of visual hull normals $\tilde{N}^{1}$ from different number of views. }
\label{fig:vhErrorDist}
\vspace{-0.3cm}
\end{figure}

To build the cost volume ($\mathtt{cv}$) for normal prediction, we sample $\phi$ uniformly from 0 to $2\pi$ and sample $\theta$ according to the visual hull normal error. In particular, we first randomly sample 100 scenes from our synthetic dataset and compute the angles between visual hull normals and ground truth normals. We set one $\theta$ value to be 0 and the other to larger than 85\% of angles between the visual hull normal $\tilde{N}^1$ and ground truth normal $\hat{N}^{1}$. The distribution of visual hull normal $\tilde{N}^1$ error for 5, 10 and 20 views are presented in Figure \ref{fig:vhErrorDist}. Table \ref{tab:sampledAngles} summarizes the configurations of $\{\theta\}$ and $\{\phi\}$ angles for different number of views.

\section{Details for Feature Mapping}
\label{sec:featureMapping}

\begin{algorithm}[t]
\footnotesize
\begin{algorithmic}
\FOR{point $\tilde{p}$ uniformly sampled from visual hull}
\STATE $\tilde{p}_{N^1}$ $\gets$ the original visual hull normal
\STATE $\tilde{p}_{M^{tr}}$ $\gets$ 1, $\tilde{p}_{I^{er}}$ $\gets$ 2, ${\tilde{p}}_{v}$ $\gets$ 0, $\tilde{p}_{c} = 0$
\ENDFOR
\FOR{view $v$ from 1 to $V$}
\FOR{point $\tilde{p}$ uniformly sampled from visual hull}
\STATE $\mathtt{isUpdate}$ $\gets$ $\mathtt{False}$
\IF{$\mathcal{V}(\tilde{p})$ = 1}
\IF {$\mathcal{S}_{v}(\tilde{p}, M^{tr}_{v})$ = 1}
\IF{$\tilde{p}_{M}^{tr}$ = 1 and $\mathtt{C}_{v}(\tilde{p}) > \tilde{p}_{c}$}
\STATE $\mathtt{isUpdate}$ $\gets$ $\mathtt{True}$
\ENDIF
\ELSE 
\IF{$\tilde{p}_{M}$ = 1}
\STATE $\mathtt{isUpdate}$ = $\mathtt{True}$
\ELSIF{$\mathcal{S}_{v}(\tilde{p}, I_{v}^{er}) < \tilde{p}_{I^{er}}$ }
\STATE $\mathtt{isUpdate}$ = $\mathtt{True}$
\ENDIF
\ENDIF
\ENDIF
\IF{$\mathtt{isUpdate}$ = $\mathtt{True}$}
\STATE $\tilde{p}_{N^{1}}$ $\gets$ $\mathcal{T}_{v}(\mathcal{S}_v(\tilde{p}, N_{v}^1) )$,~~$\tilde{p}_{M^{tr}}$ $\gets$ $\mathcal{S}_{v}(\tilde{p}, M^{tr}_{v})$
\STATE $\tilde{p}_{I^{er}}$ $\gets$ $\mathcal{S}_{v}(\tilde{p}, I^{er}_{v})$,~~$\tilde{p}_{c} = \mathcal{C}_{v}(\tilde{p})$,~~${\tilde{p}}_{v}$ $\gets$ $v$
\ENDIF
\ENDFOR
\ENDFOR
\STATE \{$f$\} $\gets$ Concatenate \{$\tilde{p}_{N^{1}}$\}, \{$\tilde{p}_{M^{tr}}$\}, \{$\tilde{p}_{I^{er}}$\}, \{$\tilde{p}_{c}$\}
\STATE \textbf{return} \{$f$\}, \{$\tilde{p}_{v}$\}
\end{algorithmic}
\caption{Mapping normals to visual hull geometry}
\label{alg:viewSelection}
\end{algorithm}

Our feature mapping method using the rendering error based view selection is summarized in Algorithm \ref{alg:viewSelection}. We first try to select the view with no total reflection as the best view $v^{*}$. If there is more than one view with no total reflection, we choose the view with the lowest rendering error. If for every view, the current point is in the region of total reflection, we choose the view whose optical center is closest to the point. Experiments in the main paper show that our rendering error based view selection ($\mathtt{RE}$) performs slightly better than average fusion ($\mathtt{AV}$) and nearest view selection ($\mathtt{NE}$) on 3D reconstruction accuracy. 

{\small
\bibliographystyle{ieee_fullname}
\bibliography{transparent}

\begin{thebibliography}{10}\itemsep=-1pt

\bibitem{OptiX}
Nvidia {O}pti{X}.
\newblock https://developer.nvidia.com/optix.

\bibitem{Atcheson2008}
Bradley Atcheson, Ivo Ihrke, Wolfgang Heidrich, Art Tevs, Derek Bradley, Marcus
  Magnor, and Hans-Peter Seidel.
\newblock Time-resolved 3{D} capture of non-stationary gas flows.
\newblock {\em ACM ToG}, 27(5):132:1--132:9, Dec. 2008.

\bibitem{BenEzra2003}
{Ben-Ezra} and {Nayar}.
\newblock What does motion reveal about transparency?
\newblock In {\em ICCV}, pages 1025--1032 vol.2, 2003.

\bibitem{besl1992method}
Paul~J. Besl and Neil~D. McKay.
\newblock Method for registration of 3-d shapes.
\newblock In {\em Sensor fusion IV: control paradigms and data structures},
  volume 1611, pages 586--606. International Society for Optics and Photonics,
  1992.

\bibitem{chang2015shapenet}
Angel~X Chang, Thomas Funkhouser, Leonidas Guibas, Pat Hanrahan, Qixing Huang,
  Zimo Li, Silvio Savarese, Manolis Savva, Shuran Song, Hao Su, et~al.
\newblock Shapenet: An information-rich 3d model repository.
\newblock {\em arXiv preprint arXiv:1512.03012}, 2015.

\bibitem{Chari2013}
Visesh Chari and Peter Sturm.
\newblock A theory of refractive photo-light-path triangulation.
\newblock In {\em CVPR}, pages 1438--1445, Washington, DC, USA, 2013. IEEE
  Computer Society.

\bibitem{Che2018}
Chengqian Che, Fujun Luan, Shuang Zhao, Kavita Bala, and Ioannis Gkioulekas.
\newblock Inverse transport networks.
\newblock {\em CoRR}, abs/1809.10820, 2018.

\bibitem{chen2019learning}
Guanying Chen, Kai Han, and Kwan-Yee~K Wong.
\newblock Learning transparent object matting.
\newblock {\em IJCV}, 127(10):1527--1544, 2019.

\bibitem{Chen2007}
T. {Chen}, H.~P.~A. {Lensch}, C. {Fuchs}, and H. {Seidel}.
\newblock Polarization and phase-shifting for 3{D} scanning of translucent
  objects.
\newblock In {\em CVPR}, pages 1--8, June 2007.

\bibitem{Chuang2000}
Yung-Yu Chuang, Douglas~E. Zongker, Joel Hindorff, Brian Curless, David~H.
  Salesin, and Richard Szeliski.
\newblock Environment matting extensions: Towards higher accuracy and real-time
  capture.
\newblock In {\em SIGGRAPH}, pages 121--130, 2000.

\bibitem{Cui2017}
Z. {Cui}, J. {Gu}, B. {Shi}, P. {Tan}, and J. {Kautz}.
\newblock Polarimetric multi-view stereo.
\newblock In {\em CVPR}, pages 369--378, July 2017.

\bibitem{Duan2015}
Qi Duan, Jianfei Cai, and Jianmin Zheng.
\newblock Compressive environment matting.
\newblock {\em Vis. Comput.}, 31(12):1587--1600, Dec. 2015.

\bibitem{gao2019deep}
Duan Gao, Xiao Li, Yue Dong, Pieter Peers, Kun Xu, and Xin Tong.
\newblock Deep inverse rendering for high-resolution svbrdf estimation from an
  arbitrary number of images.
\newblock {\em ACM Transactions on Graphics (TOG)}, 38(4):134, 2019.

\bibitem{gardner2017learning}
Marc-Andr{\'e} Gardner, Kalyan Sunkavalli, Ersin Yumer, Xiaohui Shen, Emiliano
  Gambaretto, Christian Gagn{\'e}, and Jean-Fran{\c{c}}ois Lalonde.
\newblock Learning to predict indoor illumination from a single image.
\newblock {\em arXiv preprint arXiv:1704.00090}, 2017.

\bibitem{Gregson2012}
James Gregson, Michael Krimerman, Matthias~B. Hullin, and Wolfgang Heidrich.
\newblock Stochastic tomography and its applications in 3{D} imaging of mixing
  fluids.
\newblock {\em ACM ToG}, 31(4):52:1--52:10, July 2012.

\bibitem{Han2018}
Kai Han, Kwan-Yee~K. Wong, and Miaomiao Liu.
\newblock Dense reconstruction of transparent objects by altering incident
  light paths through refraction.
\newblock {\em Int. J. Comput. Vision}, 126(5):460--475, May 2018.

\bibitem{Huynh2010}
C.~P. {Huynh}, A. {Robles-Kelly}, and E. {Hancock}.
\newblock Shape and refractive index recovery from single-view polarisation
  images.
\newblock In {\em CVPR}, pages 1229--1236, June 2010.

\bibitem{Ihrke2010}
Ivo Ihrke, Kiriakos Kutulakos, Hendrik Lensch, Marcus Magnor, and Wolfgang
  Heidrich.
\newblock Transparent and specular object reconstruction.
\newblock {\em Comput. Graph. Forum}, 29:2400--2426, 12 2010.

\bibitem{Ihrke2004}
Ivo Ihrke and Marcus Magnor.
\newblock Image-based tomographic reconstruction of flames.
\newblock In {\em Proceedings of the 2004 ACM SIGGRAPH/Eurographics Symposium
  on Computer Animation}, SCA '04, pages 365--373, Goslar Germany, Germany,
  2004. Eurographics Association.

\bibitem{Ji2013}
Y. {Ji}, J. {Ye}, and J. {Yu}.
\newblock Reconstructing gas flows using light-path approximation.
\newblock In {\em CVPR}, pages 2507--2514, June 2013.

\bibitem{kazhdan2006poisson}
Michael Kazhdan, Matthew Bolitho, and Hugues Hoppe.
\newblock Poisson surface reconstruction.
\newblock In {\em Proceedings of the fourth Eurographics symposium on Geometry
  processing}, volume~7, 2006.

\bibitem{Kim2017}
J. {Kim}, I. {Reshetouski}, and A. {Ghosh}.
\newblock Acquiring axially-symmetric transparent objects using single-view
  transmission imaging.
\newblock In {\em CVPR}, pages 1484--1492, July 2017.

\bibitem{Kutulakos2000}
Kiriakos~N. Kutulakos and Steven~M. Seitz.
\newblock A theory of shape by space carving.
\newblock {\em International Journal of Computer Vision}, 38:199--218, 2000.

\bibitem{Kutulakos2005}
K.~N. {Kutulakos} and E. {Steger}.
\newblock A theory of refractive and specular 3{D} shape by light-path
  triangulation.
\newblock In {\em ICCV}, volume~2, pages 1448--1455 Vol. 2, Oct 2005.

\bibitem{lightpath}
Kiriakos~N. Kutulakos and Eron Steger.
\newblock A theory of refractive and specular 3{D} shape by light-path
  triangulation.
\newblock {\em IJCV}, 76(1):13--29, 2008.

\bibitem{li2018differentiable}
Tzu-Mao Li, Miika Aittala, Fr\'{e}do Durand, and Jaakko Lehtinen.
\newblock Differentiable {M}onte {C}arlo ray tracing through edge sampling.
\newblock {\em ACM ToG (SIGGRAPH Asia)}, 37(6):222:1 -- 222:11, 2018.

\bibitem{li2018svbrdf}
Zhengqin Li, Zexiang Xu, Ravi Ramamoorthi, Kalyan Sunkavalli, and Manmohan
  Chandraker.
\newblock Learning to reconstruct shape and spatially-varying reflectance from
  a single image.
\newblock {\em ACM ToG (SIGGRAPH Asia)}, 37(6):269:1 -- 269:11, 2018.

\bibitem{neuralvolumes}
Stephen Lombardi, Tomas Simon, Jason Saragih, Gabriel Schwartz, Andreas
  Lehrmann, and Yaser Sheikh.
\newblock Neural volumes: {L}earning dynamic renderable volumes from images.
\newblock {\em ACM ToG (SIGGRAPH Asia)}, 38(4):65:1--65:14, 2019.

\bibitem{Matusik2002}
Wojciech Matusik, Hanspeter Pfister, Remo Ziegler, Addy Ngan, and Leonard
  McMillan.
\newblock Acquisition and rendering of transparent and refractive objects.
\newblock In {\em Eurographics Workshop on Rendering}, EGRW '02, pages
  267--278, Aire-la-Ville, Switzerland, Switzerland, 2002. Eurographics
  Association.

\bibitem{Miyazaki2005}
D. {Miyazaki} and K. {Ikeuchi}.
\newblock Inverse polarization raytracing: estimating surface shapes of
  transparent objects.
\newblock In {\em CVPR}, volume~2, pages 910--917 vol. 2, June 2005.

\bibitem{Morris2005}
N.~J.~W. {Morris} and K.~N. {Kutulakos}.
\newblock Dynamic refraction stereo.
\newblock In {\em ICCV}, volume~2, pages 1573--1580 Vol. 2, Oct 2005.

\bibitem{Morris2007}
N.~J.~W. {Morris} and K.~N. {Kutulakos}.
\newblock Reconstructing the surface of inhomogeneous transparent scenes by
  scatter-trace photography.
\newblock In {\em ICCV}, pages 1--8, Oct 2007.

\bibitem{Peers2003}
Pieter Peers and Philip Dutr{\'e}.
\newblock Wavelet environment matting.
\newblock In {\em Proceedings of the 14th Eurographics Workshop on Rendering},
  EGRW '03, pages 157--166, Aire-la-Ville, Switzerland, Switzerland, 2003.
  Eurographics Association.

\bibitem{qi2017pointnet++}
Charles~Ruizhongtai Qi, Li Yi, Hao Su, and Leonidas~J Guibas.
\newblock Pointnet++: {D}eep hierarchical feature learning on point sets in a
  metric space.
\newblock In {\em Advances in neural information processing systems}, pages
  5099--5108, 2017.

\bibitem{Qian2015}
Y. {Qian}, M. {Gong}, and Y. {Yang}.
\newblock Frequency-based environment matting by compressive sensing.
\newblock In {\em ICCV}, pages 3532--3540, Dec 2015.

\bibitem{Qian2017}
Y. {Qian}, M. {Gong}, and Y. {Yang}.
\newblock Stereo-based 3d reconstruction of dynamic fluid surfaces by global
  optimization.
\newblock In {\em CVPR}, pages 6650--6659, July 2017.

\bibitem{Qian2016}
Yiming Qian, Minglun Gong, and Yee-Hong Yang.
\newblock 3d reconstruction of transparent objects with position-normal
  consistency.
\newblock In {\em CVPR}, pages 4369--4377, 06 2016.

\bibitem{schoenberger2016sfm}
Johannes~Lutz Sch\"{o}nberger and Jan-Michael Frahm.
\newblock Structure-from-motion revisited.
\newblock In {\em Conference on Computer Vision and Pattern Recognition
  (CVPR)}, 2016.

\bibitem{Seitz2006}
S.~M. {Seitz}, B. {Curless}, J. {Diebel}, D. {Scharstein}, and R. {Szeliski}.
\newblock A comparison and evaluation of multi-view stereo reconstruction
  algorithms.
\newblock In {\em CVPR}, volume~1, pages 519--528, June 2006.

\bibitem{Shan2012}
Qi Shan, Sameer Agarwal, and Brian Curless.
\newblock Refractive height fields from single and multiple images.
\newblock In {\em CVPR}, pages 286--293, 06 2012.

\bibitem{stets2019single}
Jonathan Stets, Zhengqin Li, Jeppe~Revall Frisvad, and Manmohan Chandraker.
\newblock Single-shot analysis of refractive shape using convolutional neural
  networks.
\newblock In {\em 2019 IEEE Winter Conference on Applications of Computer
  Vision (WACV)}, pages 995--1003. IEEE, 2019.

\bibitem{jonathanacquisition}
Jonathan~Dyssel Stets, Alessandro~Dal Corso, Jannik~Boll Nielsen,
  Rasmus~Ahrenkiel Lyngby, Sebastian Hoppe~Nesgaard Jensen, Jakob Wilm,
  Mads~Brix Doest, Carsten Gundlach, Eythor~Runar Eiriksson, Knut Conradsen,
  Anders~Bjorholm Dahl, Jakob~Andreas B{\ae}rentzen, Jeppe~Revall Frisvad, and
  Henrik Aan{\ae}s.
\newblock Scene reassembly after multimodal digitization and pipeline
  evaluation using photorealistic rendering.
\newblock {\em Appl. Optics}, 56(27):7679--7690, 2017.

\bibitem{Tanaka2016}
K. {Tanaka}, Y. {Mukaigawa}, H. {Kubo}, Y. {Matsushita}, and Y. {Yagi}.
\newblock Recovering transparent shape from time-of-flight distortion.
\newblock In {\em CVPR}, pages 4387--4395, June 2016.

\bibitem{Trifonov2006}
Borislav Trifonov, Derek Bradley, and Wolfgang Heidrich.
\newblock Tomographic reconstruction of transparent objects.
\newblock In {\em ACM SIGGRAPH 2006 Sketches}, SIGGRAPH, New York, NY, USA,
  2006. ACM.

\bibitem{Tsai2015}
C. {Tsai}, A. {Veeraraghavan}, and A.~C. {Sankaranarayanan}.
\newblock What does a single light-ray reveal about a transparent object?
\newblock In {\em ICIP}, pages 606--610, Sep. 2015.

\bibitem{Wetzstein2011}
G. {Wetzstein}, D. {Roodnick}, W. {Heidrich}, and R. {Raskar}.
\newblock Refractive shape from light field distortion.
\newblock In {\em ICCV}, pages 1180--1186, Nov 2011.

\bibitem{Wexler2002}
Yonatan Wexler, Andrew Fitzgibbon, and Andrew Zisserman.
\newblock Image-based environment matting.
\newblock In {\em CVPR}, pages 279--290, 01 2002.

\bibitem{wu2018full}
Bojian Wu, Yang Zhou, Yiming Qian, Minglun Cong, and Hui Huang.
\newblock Full 3d reconstruction of transparent objects.
\newblock {\em ACM ToG}, 37(4):103:1--103:11, July 2018.

\bibitem{Wu2015}
Zhaohui Wu, Zhong Zhou, Delei Tian, and Wei Wu.
\newblock Reconstruction of three-dimensional flame with color temperature.
\newblock {\em Vis. Comput.}, 31(5):613--625, May 2015.

\bibitem{xu2018deep}
Zexiang Xu, Kalyan Sunkavalli, Sunil Hadap, and Ravi Ramamoorthi.
\newblock Deep image-based relighting from optimal sparse samples.
\newblock {\em ACM Transactions on Graphics (TOG)}, 37(4):126, 2018.

\bibitem{Yao2019}
Yao Yao, Zixin Luo, Shiwei Li, Tianwei Shen, Tian Fang, and Long Quan.
\newblock Recurrent {M}{V}{S}{N}et for high-resolution multi-view stereo depth
  inference.
\newblock In {\em CVPR}, June 2019.

\bibitem{Yeung2011}
S. {Yeung}, T. {Wu}, C. {Tang}, T.~F. {Chan}, and S. {Osher}.
\newblock Adequate reconstruction of transparent objects on a shoestring
  budget.
\newblock In {\em CVPR}, pages 2513--2520, June 2011.

\bibitem{Yeung2014}
S. {Yeung}, T. {Wu}, C. {Tang}, T.~F. {Chan}, and S.~J. {Osher}.
\newblock Normal estimation of a transparent object using a video.
\newblock {\em PAMI}, 37(4):890--897, April 2015.

\bibitem{yeungetal}
Sai-Kit Yeung, Chi-Keung Tang, Michael~S. Brown, and Sing~Bing Kang.
\newblock Matting and compositing of transparent and refractive objects.
\newblock {\em ACM ToG (SIGGRAPH)}, 30(1):2:1--2:13, 2011.

\bibitem{zhang2019DTRT}
Cheng Zhang, Lifan Wu, Changxi Zheng, Ioannis Gkioulekas, Ravi Ramamoorthi, and
  Shuang Zhao.
\newblock A differential theory of radiative transfer.
\newblock {\em ACM Trans. Graph.}, 38(6), 2019.

\bibitem{Zhang2014}
Mingjie Zhang, Xing Lin, Mohit Gupta, Jinli Suo, and Qionghai Dai.
\newblock Recovering scene geometry under wavy fluid via distortion and defocus
  analysis.
\newblock In {\em ECCV}, volume 8693, pages 234--250, 09 2014.

\bibitem{Zhu2004}
Jiayuan Zhu and Yee-Hong Yang.
\newblock Frequency-based environment matting.
\newblock In {\em Pacific Graphics}, pages 402--410, 2004.

\bibitem{Zongker1999}
Douglas~E. Zongker, Dawn~M. Werner, Brian Curless, and David~H. Salesin.
\newblock Environment matting and compositing.
\newblock In {\em Proceedings of the 26th Annual Conference on Computer
  Graphics and Interactive Techniques}, SIGGRAPH '99, pages 205--214, New York,
  NY, USA, 1999. ACM Press/Addison-Wesley Publishing Co.

\bibitem{Zuo2015}
X. {Zuo}, C. {Du}, S. {Wang}, J. {Zheng}, and R. {Yang}.
\newblock Interactive visual hull refinement for specular and transparent
  object surface reconstruction.
\newblock In {\em ICCV}, pages 2237--2245, Dec 2015.

\end{thebibliography}
}

\end{document}